%% file: Zoe.tex
\documentclass[journal,twoside,web]{ieeecolor}
\PassOptionsToPackage{table}{xcolor}
\usepackage{jsen}
\usepackage{cite}
\usepackage{amsmath,amssymb,amsfonts}
\usepackage{algorithmic}
\usepackage{graphicx}
\usepackage{textcomp}
\usepackage{wrapfig}
\usepackage{booktabs}
\usepackage{algorithm}
\usepackage{url}
\usepackage{multido}
\usepackage{tabularx} % Required for foldable/wrapped text

\def\BibTeX{{\rm B\kern-.05em{\sc i\kern-.025em b}\kern-.08em
    T\kern-.1667em\lower.7ex\hbox{E}\kern-.125emX}}
\definecolor{abstractbg}{rgb}{0.89804,0.94510,0.83137}
\begin{document}
\title{Calibration-free IMU-based Relative Orientation Estimation under Dynamic Human Motion}
\author{Ruiyuan Li,  Hassan Osman,  Ajay Seth, and Manon Kok 
\thanks{Manuscript received XX XX, 2026; revised XX XX, 2026. (Corresponding author: Ruiyuan Li.)}
\thanks{Ruiyuan Li and Manon Kok are with the Delft Center for Systems and Control, Delft University of Technology, 2628 CD Delft, the Netherlands (e-mail: R.Li-6@tudelft.nl; M.Kok-1@tudelft.nl).}
\thanks{Hassan Osman and Ajay Seth are with the Department of Biomechanical Engineering, Delft University of Technology, 2628 CD Delft, the Netherlands (e-mail: H.Osman@tudelft.nl; A.Seth@tudelft.nl).}
}
% \thanks{}

\IEEEtitleabstractindextext{%
\fcolorbox{abstractbg}{abstractbg}{%
\begin{minipage}{\textwidth}%、
\begin{wrapfigure}{R}{0.48\textwidth} % r = right side, width = 40% of text width
   \vspace{-10pt}
    \centering
    \includegraphics[width=0.95\linewidth,trim={0 0 0 0}]{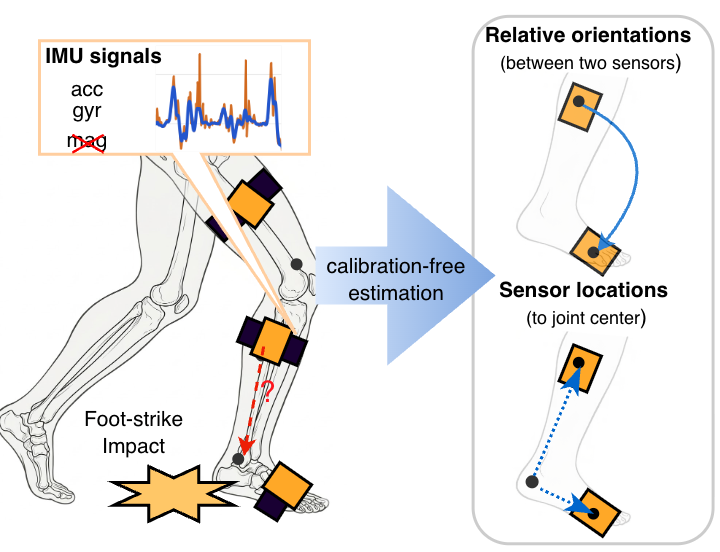}
    \caption{ Calibration-free relative orientation and sensor location estimation from magnetometer-free IMU measurements.
    \label{fig:sys_diag}}
    \vspace{-15pt}
\end{wrapfigure}
\begin{abstract}
% Recent work on magnetometer-free IMU based motion estimation has shown the ability to implement drift-free estimation. However, it requires either pre-processing to know IMU location on segments or constraints to 1-d joint motion (for example, knee).  Recent work achieves 3d joint relative IMU orientation tracking by estimating joint location online. However, it requires one IMU orientation anchor and bypasses the real word dataset. we proposed the unified 12-state extended Kalman filter (EKF) with optional Low-pass filter (LPF) on measurement update estimating relative joint orientation and IMU-to-joint location simultaneously from raw dual-IMU accelerations and angular velocities. On both the simulation and real datasets, we demonstrated that the proposed method does not degrade the performance of the relative orientation tracking, requiring neither pre-processing nor a global orientation anchor. The simulation across 490 trials sweeping 7 different std of the randomly generated angular velocity signals and joint center acceleration; the orientation estimation accuracy and sensor location accuracy is governed by gravity-free 
Recent magnetometer-free inertial measurement unit (IMU) methods can reduce drift in joint motion estimation using rigid-body constraints, but they typically require pre-calibrated sensor locations, assumptions of specific joint models (e.g., hinge joints) or limited motion types,  or an external reference for global orientation.
Furthermore, challenges related to high-frequency impacts (e.g. at foot strikes during gait) and numerical differentiation of noisy measurements have not been explicitly addressed.
We propose an extended Kalman filter that estimates relative IMU orientation and sensor location simultaneously, requiring neither offline sensor location calibration nor a global orientation anchor. 
 A systematic simulation study shows that the quality of the relative orientation and the sensor location estimates depends on different motion-intensity sources: the joint center acceleration and the relative angular velocity, respectively.
 A low-pass filter (LPF) is used to mitigate real-signal outliers before applying the rigid-body constraint. 
When applied to the OpenSense dataset with real IMU measurements across four subjects with three lower-body joint types during walking and complex tasks including various motion types, the proposed method achieves an overall relative orientation error of 2.61$^\circ$, compared with 10.40$^\circ$ for a baseline using offline-estimated sensor locations and 7.43$^\circ$ for a baseline using marker-derived sensor locations.   These results show that the proposed calibration-free estimation framework handles real IMU data effectively and achieves accurate relative orientation without offline sensor location calibration.

\end{abstract}
\end{minipage}}}
\maketitle
\setcounter{figure}{1} % Force the next figure to be Figure 2
\input{chapters/introduction}

\input{chapters/sec_II_temp}

\input{chapters/sec_III_temp}

\input{chapters/sec_IV}

\input{chapters/sec_V_A}

% \input{chapters/sec_V_D}
\input{chapters/conclusion}

% \section*{Acknowledgment}

 \bibliographystyle{IEEEtran}
  \bibliography{reference} 
  \vspace{-1cm}
\begin{IEEEbiography}[{\includegraphics[width=1in,clip,trim={170 70 170 0}]{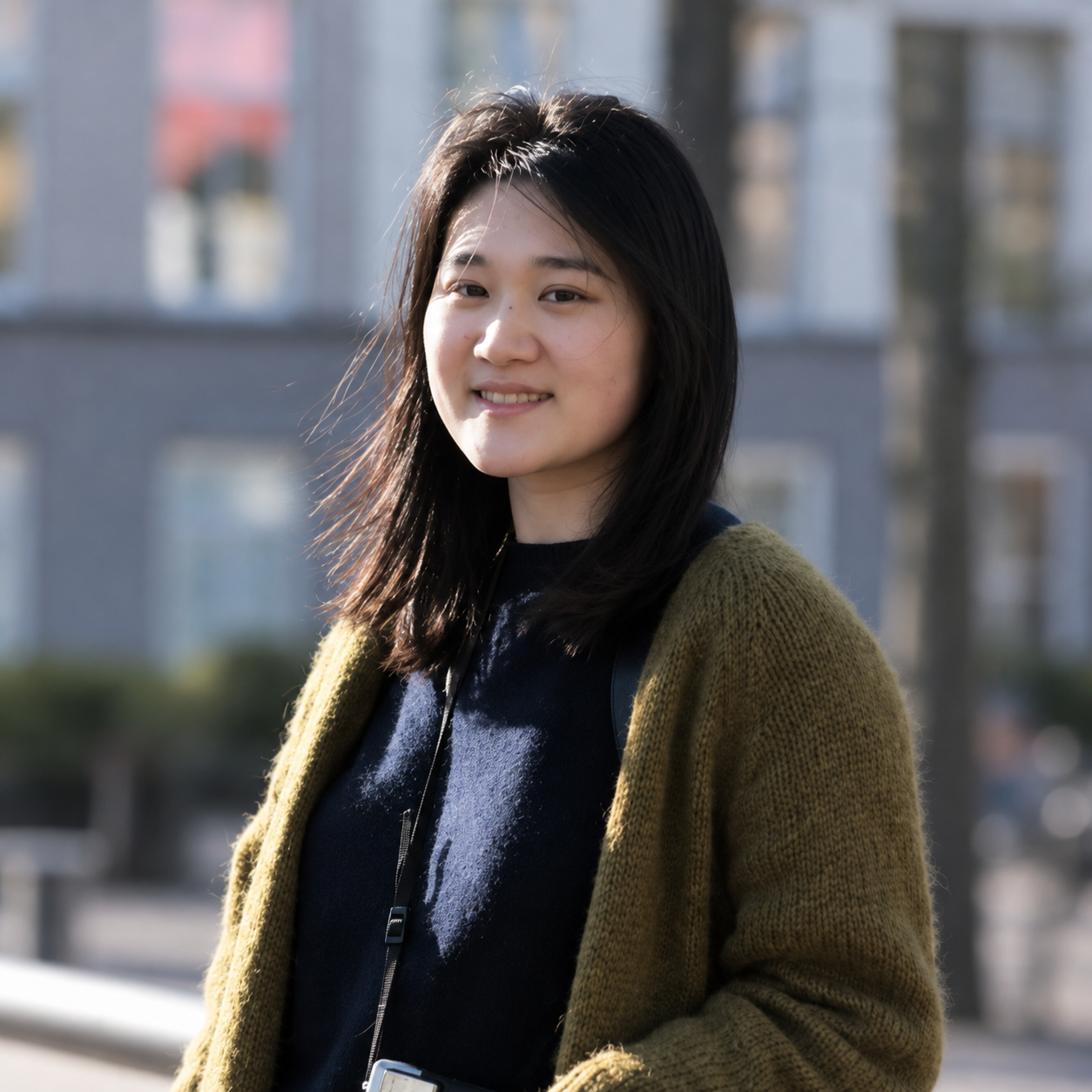}}]{Ruiyuan Li}
received the M.Sc. degree in computer science from the University of Twente, Enschede, the Netherlands, in 2022. She is currently pursuing the Ph.D. degree at Delft University of Technology, the Netherlands. Her research interests include human motion understanding, physics-informed machine learning, and multimodal perception.

\end{IEEEbiography}
\vspace{-0.8cm}
% --- Author 2 ---
\begin{IEEEbiography}[{\includegraphics[width=1in,height=1.25in,clip,keepaspectratio]{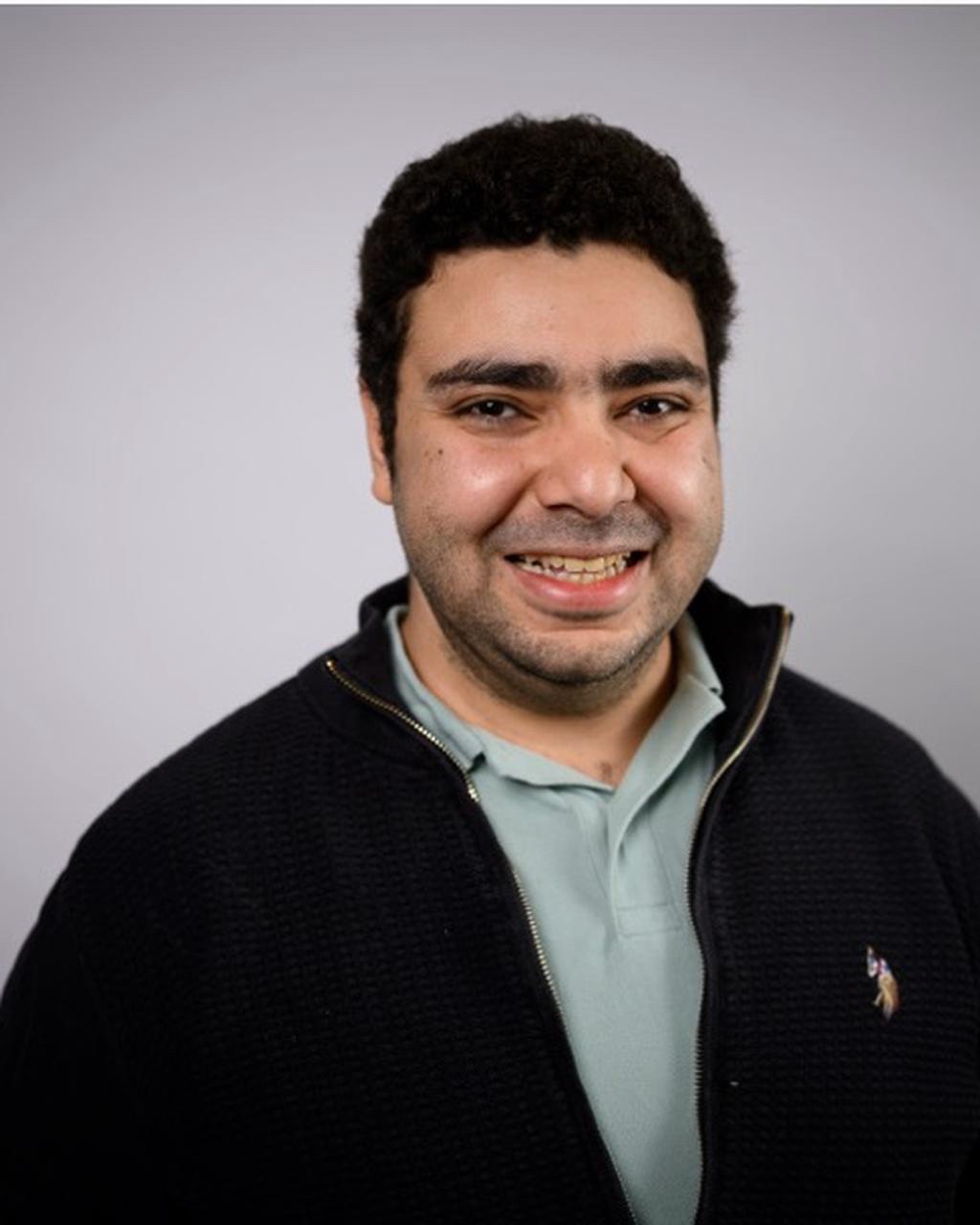}}]{Hassan Osman}
received the B.Sc. and M.Sc. degrees in Mechatronics Engineering from Ain Shams University, Cairo, Egypt. He is currently pursuing the Ph.D. degree at Delft University of Technology, The Netherlands, with a focus on human motion estimation using wearable devices.
\end{IEEEbiography}
\vspace{-0.8cm}
% % --- Author 3 ---
\begin{IEEEbiography}[{\includegraphics[width=1in,height=1.25in,clip,keepaspectratio]{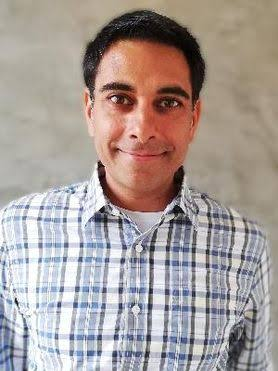}}]{Ajay Seth}
 received the B.Sc. and M.Sc. degrees in Systems Design Engineering from the University of Waterloo, Canada, and the Ph.D. degree in Biomedical Engineering from The University of Texas at Austin, USA.
He is currently an Associate Professor of Biomechanical Engineering at Delft University of Technology, the Netherlands. He directs the computational biomechanics laboratory with a focus on engineering open-source neuromusculoskeletal modeling and simulation tools for human movement analysis.
\end{IEEEbiography}
\vspace{-0.8cm}
% % --- Author 4 ---
\begin{IEEEbiography}[{\includegraphics[width=1in,height=1.25in,clip,keepaspectratio]{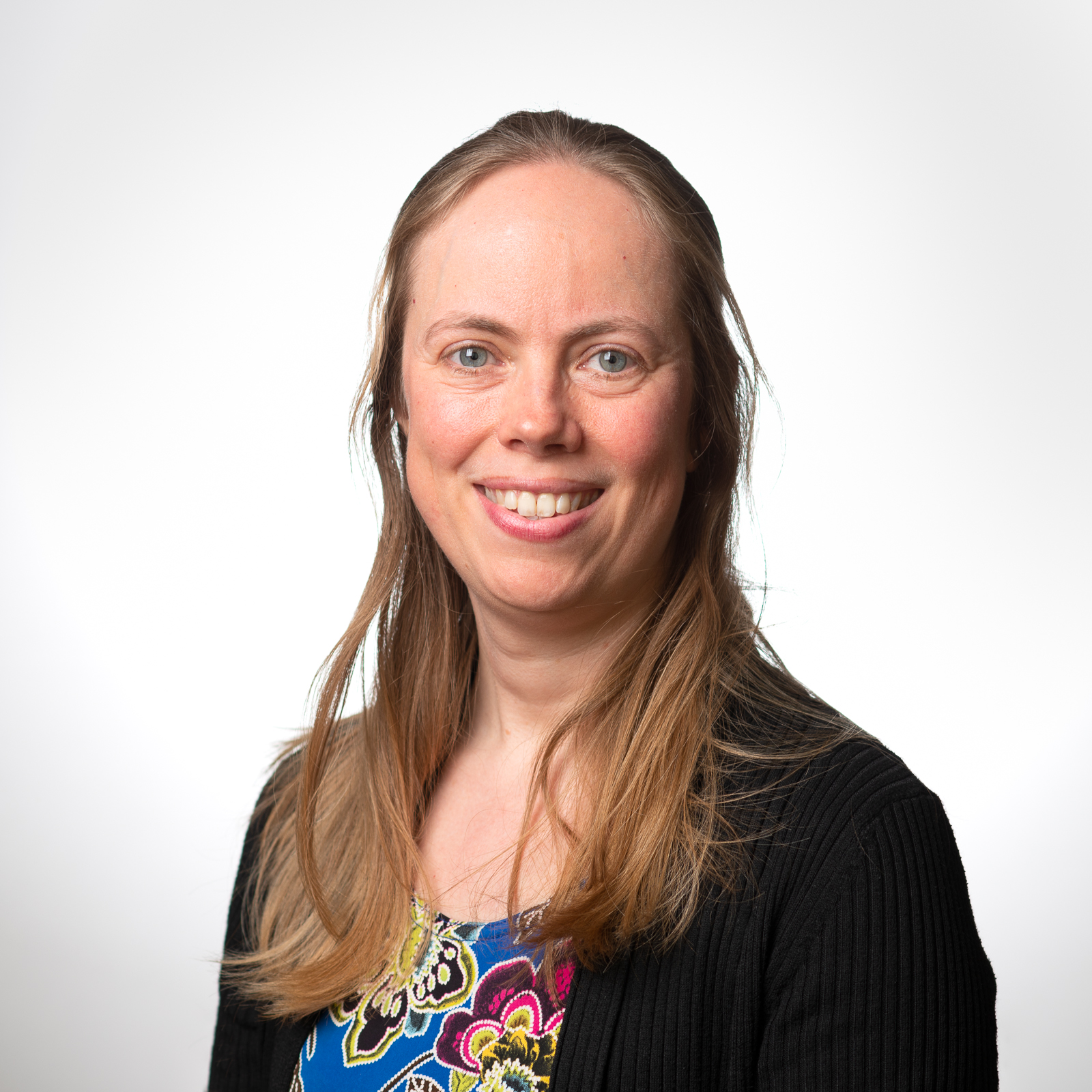}}]{Manon Kok}
received M.Sc.\ degrees in philosophy of science, technology, and society, and in applied physics from the University of Twente, the Netherlands, and the Ph.D.\ degree in automatic control from Linköping University, Sweden. She is currently an Associate Professor at Delft University of Technology, the Netherlands. Her research interests include probabilistic inference for sensor fusion, signal processing, and machine learning.
\end{IEEEbiography}

\end{document}

%% file: chapters/introduction.tex
% ============================================================
% Section I: Introduction (with related work inline)
% IEEE Sensors Journal — Calibration-free IMU Joint Orientation
% Target: ~1 column (tight; 8-page limit applies)
% Notation: joint-center location vectors = \tilde{\mathbf{r}}_i^{S_i}
%   (consistent with algorithm_method.tex)
% TODO: make all expressions consistent across sections before submission
% ============================================================

\section{Introduction}
\label{sec:introduction}

Accurate joint kinematics estimation is fundamental to human motion tracking and reconstruction, with  applications in rehabilitation,  sports, and entertainment \cite{lam2023systematic, menache:2000, suoTL:2024}. Inertial measurement units (IMUs) are a practical option for human motion estimation, as they are inexpensive, portable and lightweight, which allows their use outside the laboratory  \cite{lopez2016wearable, picerno201725, kok2026inertial}. Joint kinematics can be estimated from single IMU (global) orientation estimates, which are  commonly obtained by fusing gyroscope, accelerometer, and magnetometer measurements~\cite{madgwick2011estimation, laidig2023vqf}. However, sensor errors, e.g.\ due to indoor magnetic disturbances corrupting the magnetometer measurements, lead to erroneous orientation estimates.  
Since joint kinematics such as joint angles are described by the relative orientation between adjacent body segments, typically the relative orientation is of most interest. Relative orientation between adjacent sensors can be estimated without the use of magnetometers~\cite{kok2026inertial}, which is the focus of this work. Note that our focus is on relative sensor orientation, and sensor-to-segment calibration to obtain relative segment orientation is therefore outside the scope.
% , provided that the  segment-to-sensor calibration is known.
%In this work, we focus on 

Existing methods for relative orientation estimation typically rely on assumptions regarding joint type~\cite{dejnabadi2005new, dorschky2019estimation}, offline calibration~\cite{weygers2020drift}, or have primarily been validated on simulated IMU measurements~\cite{lorenz2025minjointtracker} or real walking motions~\cite{weygers2020drift, taetz2025jointtracker}, limiting their practical applicability. 
In this work, we enable magnetometer-free and calibration-free 3D joint kinematic estimation using real IMU measurements during diverse human motions.

To this end, we extend the method of~\cite{weygers2020drift} to jointly estimate the relative sensor orientation and sensor locations online using only acceleration and angular velocity measurements, eliminating the need for offline sensor location calibration.
A systematic simulation study shows that, under sufficient joint center acceleration intensity, online sensor location estimation does not degrade orientation estimation accuracy. It further shows that orientation and sensor location estimation are primarily governed by joint center acceleration and relative angular velocity, respectively. 
To improve reliability on real IMU measurements, we mitigate high-frequency measurement outliers caused by foot-strike impacts in the acceleration and numerical differentiation of angular velocity using a low-pass filter (LPF).
Finally, the proposed framework is extensively validated on the OpenSense dataset~\cite{al2022opensense}, containing walking and diverse lower-body motions.

\section{related work}

Recent advances in inertial motion capture have moved toward magnetometer-free methods \cite{kok2026inertial}. Some magnetometer-free methods focus on lower-dimensional or planar joint models~\cite{dejnabadi2005new, dorschky2019estimation}, whereas~\cite{weygers2020drift} incorporated a joint-connection model, modeling the connection of adjacent segments by a 3D joint.  This enabled long-term, drift-free relative IMU orientation estimation using an extended Kalman filter (EKF) without assuming specific joint or motion types. 
The approach relies on knowledge of where the sensors are placed on the body, which was determined in~\cite{weygers2020drift} using a separate offline calibration step~\cite{seel2012joint}. In contrast, our method estimates the sensor locations online, jointly with the relative IMU orientation, removing the need for this offline calibration.

%Recent EKF-based approaches have incorporated online IMU location estimation in dynamic human motion estimation. Lorenz et al.~\cite{lorenz2025minjointtracker} proposed a recursive Bayesian filter that jointly estimates IMU locations and orientations online for a lower-body kinematic chain. 

Joint relative IMU orientation and sensor location estimation has also been studied in~\cite{lorenz2025minjointtracker}. Using the orientation of one IMU as the anchor, they estimate the global orientation of a lower-body kinematic chain and validate their method on re-simulated IMU signals synthesized from marker trajectories to reduce artifacts present in real IMU measurements. Similarly, \cite{taetz2025jointtracker} jointly estimates each IMU's global orientation and sensor location using multiple kinematic constraints, including the joint connection model, and uses magnetometer measurements for initialization.
The results are evaluated on human gait motion and humanoid robot data. Although these methods enable online sensor location estimation, the effects of artifacts in real IMU measurements have not been systematically addressed.
% they do not systematically handle artifacts in real IMU measurements that violate  standard Gaussian measurement noise assumptions.

Such artifacts result from impact-related transients~\cite{Mitschke2017,Fan2025}. Numerical differentiation required to evaluate the joint-connection model is highly sensitive to these artifacts~\cite{winter2009biomechanics}, potentially violating the Gaussian measurement noise assumption.
% In real human-motion data, however, Gaussian measurement-noise assumptions can be violated by impact-related transients~\cite{Mitschke2017,Fan2025} and by artifacts introduced through the numerical differentiation required to evaluate the joint-connection model~\cite{Winter1990}. 
Low-pass filtering (LPF) has been shown to reduce IMU measurement errors during impulsive motion~\cite{Rong2024}. We therefore incorporate LPF into the proposed method and evaluate its effect on the joint-connection model using real IMU measurements. Inspired by the observability analysis from~\cite{kok2022observability}, we further study how estimation performance varies with motion intensity, first systematically in simulation and then using real human-motion data. Additionally, we evaluate the proposed method on a broader range of human motions than the walking motions considered previously~\cite{taetz2025jointtracker,lorenz2025minjointtracker,weygers2020drift}.

%% file: chapters/sec_II_temp.tex
\section{Modeling}
\label{sec:modeling}
\begin{figure}[h]
\centering
\includegraphics[width=1\linewidth]{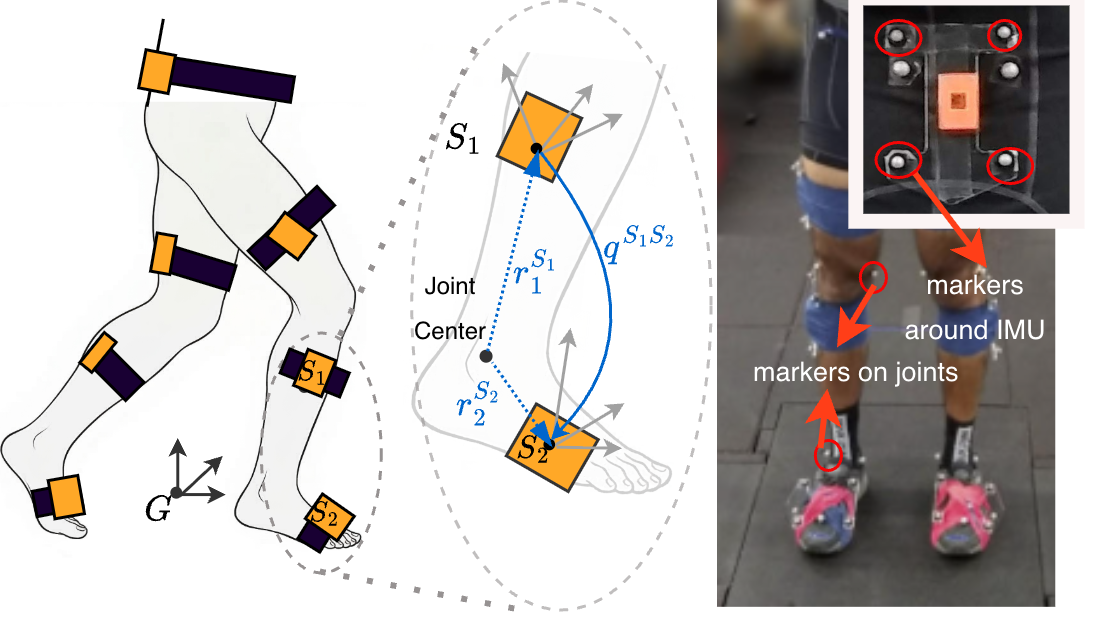}
\caption{Left: Graphical illustration of the relative orientation $q^{S_1S_2}$ and sensor location $r_i^{S_i}$. Right: Example of a subject equipped with IMUs and markers in the OpenSense dataset~\cite{al2022opensense}.}
\label{fig:sensor_diagram}
\end{figure}

We consider two rigid body segments connected by a 3D  joint that encompasses 1D and 2D joints and motions. To each segment, an IMU is attached at an arbitrary location and  orientation. Let $G$ denote a global Earth-fixed frame, $S_i$ ($i\in\{1,2\}$) denote the sensor frame of sensor $i$, and let us assume that sensor $i$ is rigidly attached to segment $i$. We are interested in the \emph{relative} orientation $q^{S_1S_2}$, where the superscripts explicitly indicate that this is the rotation from sensor frame $S_2$ to $S_1$, as illustrated in Fig.~\ref{fig:sensor_diagram}. To determine $q_t^{S_1S_2}$, we first estimate the individual \emph{absolute} sensor orientations $q_t^{GS_i}$. 

The assumption of the two body segments being connected is considered a measurement model. It requires knowledge about how the sensors are placed on the body segment. More specifically, we denote the sensor location vector from the joint center to each sensor $i$ by $r_i^{S_i}$, expressed in sensor frame $S_i$. To enable calibration-free estimation, we estimate $r_i^{S_i}$ jointly with the sensor orientations, for $i=1,2$.

% The location of the two sensors with respect to the joint center connecting sensors 1 and 2 are denoted by $r_1^{S_1}$ and $r_2^{S_2}$. 
% We assume the sensor orientations and the location vectors $r_i^{S_i}, i = 1, 2$ to be unknown.

% The vectors $r_i^{S_i}$ are \emph{unknown} and estimated jointly with the orientation $q_t^{GS_i}$ using an error-state-space model following \cite{kok2017using}, and the formal state vector is defined in Section~\ref{sec:algorithm}. 

% The discrete time index is $t = 1,\ldots,N$, where $N$ is the number of samples.

%--------------------------------------------------------------
\subsection{Dynamic Model}
%--------------------------------------------------------------

The dynamics of the orientation for each sensor $i = 1, 2$ is modeled as
\begin{equation}
    {q}_{t+1}^{GS_i} = {q}_t^{GS_i} \odot\exp_\text{q}\!\left( \frac{T}{2} \bar{{y}}_{\omega,t}^{S_i} \right),
    \label{eq:quat_prop}
\end{equation}
where $q^{GS_i}_{t}$ is the quaternion representing the absolute sensor orientation at time $t$, $\odot$ denotes quaternion multiplication, $\exp_q(\cdot)$ the quaternion exponential map, and $\bar{{v}} \triangleq [0,\, {v}^\top]^\top$ denotes the quaternion representation of a vector ${v} \in \mathbb{R}^3$ \cite{kok2017using}. Furthermore,~$T$ denotes the sampling period and ${y}_{\omega,t}^{S_i}$ the gyroscope measurement of sensor $i$, which is modeled as
\begin{equation}
    {y}_{\omega,t}^{S_i} = \omega_t^{S_i} + {e}_{\omega,t}^{S_i}.
    \label{eq:gyr_meas}
\end{equation}
Here, $\omega_t^{S_i} \in \mathbb{R}^3$ is the angular velocity of segment $i$, and ${e}_{\omega,t}^{S_i} \sim \mathcal{N}({0}, {\Sigma}_\omega)$ is zero-mean Gaussian white noise with covariance ${\Sigma}_\omega$.
% We assume the sensor-to-segment location to be nearly constant and therefore model the dynamics of ${{r}}_{t}^{S_i}$ is modeled as
We assume that each sensor location ${{r}}_i^{S_i}$ is nearly constant and model its dynamics as
\begin{equation}
    {{r}}_{i,t}^{S_i} = {{r}}_{i,t-1}^{S_i} + {e}_{r,t}^{S_i},
    \label{eq:r_rw}
\end{equation}
where ${e}_{r,t}^{S_i} \sim \mathcal{N}({0}, {\Sigma}_r)$. 
% A small ${\Sigma}_r$ suggests that the sensor location is constant.

%--------------------------------------------------------------
\subsection{Measurement Model}
%--------------------------------------------------------------

% \subsubsection{Kinematic Constraint}

The joint-center acceleration ${a}_{jc,t}^G$ in the  global frame can be computed equivalently from quantities obtained from sensor  $1$ and from sensor  $2$. Hence, 
% In the connected rigid body system we considered, since both sensor frames are rigidly attached to the same joint, the joint-center acceleration expressed in the global frame must be identical for both segments:
\begin{equation}
    {R}_t^{GS_1} {a}_{jc,t}^{S_1} = {R}_t^{GS_2} {a}_{jc,t}^{S_2} + {e}_{\text{link},t},
    \label{eq:constraint_phys}    
\end{equation}
where ${R}_t^{GS_i} \in SO(3)$ denotes the rotation matrix representation of the quaternion $q_t^{GS_i}$. Gaussian noise ${e}_{\text{link},t} \sim \mathcal{N}({0}, {\Sigma}_{\text{link}})$ denotes the model mismatch, and ${a}_{jc,t}^{S_i} \in \mathbb{R}^3$ is the joint-center acceleration modeled as
% .  Since $a_{jc,t}^{S_i}$ is not directly measured, we model the rigid-body kinematics as
\begin{equation}
    {a}_{jc,t}^{S_i} = {a}_{t}^{S_i} - (\left[{\omega}_{t}^{S_i} \times\right]^2 + \left[\dot{\omega}_{t}^{S_i} \times\right]) {r}_i^{S_i},
    ~\label{eq:ajc}
\end{equation} 
where $[\cdot\times]$ denotes the cross-product matrix and $\dot{\omega}_{t}^{S_i}$ the angular acceleration. The angular velocity $\omega_t^{S_i}$ is obtained from the gyroscope measurement model~\eqref{eq:gyr_meas}, and $\dot{\omega}_t^{S_i}$ is obtained through numerical differentiation of the gyroscope measurements. The sensor acceleration $a_t^{S_i}$ denotes the ideal accelerometer output (specific force), and is modeled as
% Typically,   ${\omega}_{t}^{S_i}$ is approximated using ${y}_{\omega,t}^{S_i}$ modeled by~\eqref{eq:gyr_meas}, and $\dot{\omega}_t^{S_i}$ is calculated using 5-point numerical differentiation method applied to the gyroscope signal ${y}_{\omega}^{S_i}$. The sensor  acceleration ${a}_{t}^{S_i} $ is approximated by the measurement ${y}_{a,t}^{S_i}$ modeled as
\begin{equation}
    {y}_{a,t}^{S_i} = {a}_t^{S_i} + {e}_{a,t}^{S_i},
    \label{eq:acc_meas}
\end{equation}
where ${y}_{a,t}^{S_i}$ denotes the accelerometer measurements and $e_{a,t}^{S_i}$ the measurement noise.

%% file: chapters/sec_III_temp.tex
\section{Method}
\label{sec:algorithm}

Extending the filtering method from~\cite{weygers2020drift}, we propose an EKF that jointly estimates relative orientation $q^{S_1S_2}$ and sensor location vectors $r_i^{S_i}$, enabling calibration-free joint kinematic estimation.
% Following the dynamic model \eqref{eq:quat_prop}--\eqref{eq:r_rw} 
% Following~\cite{kok2017using}, we estimate $q^{GS_i}$, $i=1,2$ using an orientation-deviation formulation. 
Following the filtering formulation in~\cite{weygers2020drift}, we first estimate the absolute orientations $q^{GS_i}$, $i=1,2$, parametrized using the orientation-deviation formulation in~\cite{kok2017using}, from which we determine  $q^{S_1S_2}$. Hence, the state is defined as
\begin{equation}
    {x}_t = \begin{bmatrix}
      {{\eta}}_{1,t}^G ,
     {{\eta}}_{2,t}^G ,
    {{r}}_{1,t}^{S_1} ,
      {{r}}_{2,t}^{S_2}
    \end{bmatrix}^\top \in \mathbb{R}^{12},
    \label{eq:state}
\end{equation}
% where $ {{r}}_{i,t}^{S_i}$, i = 1, 2, are the sensor location vectors and ${{\eta}}_{i,t}^G \in \mathbb{R}^3$ the orientation deviation from a linearization point represented by the quaternion ${\hat{q}}_{t|t}^{GS_i}$.
% where ${{r}}_{i,t}^{S_i}$, $i = 1, 2$, are the sensor location vectors, 
where ${{\eta}}_{i,t}^G \in \mathbb{R}^3$ are orientation deviations from the current orientation estimate $\hat{{q}}_{t}^{GS_i}$. 

% Here, $t|t$ and $t|t-1$ denote the posterior and prior estimates at time $t$, respectively.

The dynamic model \eqref{eq:quat_prop}--\eqref{eq:r_rw} is used in the time update of the EKF (steps~3--7 in Alg.~\ref{alg:ekf}). The time update sets $\hat{\eta}_{i,t|t-1}^G$ to zero and directly updates the linearization point $\hat{{q}}_{t|t-1}^{GS_i}$ using \eqref{eq:quat_prop}. 
%In each iteration, we reset $\hat{\eta}_{i,t|t-1}^G$ to zero at the new linearization point, and this update  propagates $\hat{{q}}_{t|t-1}^{GS_i}$ via the gyroscope 
% and keeps $\hat{{r}}_{i,t|t-1}^{S_i}$ constant
 %and keeps $\hat{r}_{i,t|t-1}^{S_i}$ unchanged. 
 Here, $t|t$ and $t|t-1$ denote the posterior and prior estimates at time $t$, respectively. The process noise covariance~$Q$ in step~7 of Alg.~\ref{alg:ekf} is block-diagonal, reflecting independent process noise on the orientation deviations and the sensor location vectors as
\begin{equation}
Q =\mathrm{blkdiag}\!\left(
                            {\Sigma}_\omega,\;
                           {\Sigma}_\omega,\;
                            {\Sigma}_r,\;
                            {\Sigma}_r
                        \right).
    \label{eq:Q}
\end{equation}
Here, ${\Sigma}_\omega = \sigma_\omega^2\,\mathcal{I}_3$ is the gyroscope process noise covariance, and $\mathcal{I}_n$ denotes the $n \times n$ identity matrix. Furthermore, ${\Sigma}_r = \sigma_r^2\,\mathcal{I}_3$ is chosen such that it encodes the assumption that $r_i^{S_i}$ is (approximately) constant.
% , with $\sigma_r^2 \ll \sigma_\omega^2$. 

\begin{algorithm}[h]
\caption{\label{alg:ekf} Calibration-free Relative  Orientation Estimation}
\textbf{Input:} Raw accelerometer $\{ {y}_{a,t}^{S_i} \}_{t=1}^{N}$ and gyroscope $\{ {y}_{\omega,t}^{S_i} \}_{t=1}^{N}$ data, initial quaternion $\hat{{q}}_{1|1}^{GS_i}$ and  location vector estimates $\hat{r}_{i,1|1}^{S_i}$, $i=1,2$, and  initial state covariance $P_{1|1}$, process noise covariance $Q$, measurement noise covariance $R$.
% \textbf{Input:} Filtered accelerometer $\{ \breve{{y}}_{a,t}^{S_i} \}_{t=1}^{N}$;
% filtered gyroscope copy for $\dot{{\omega}}$ only $\{ \breve{{y}}_{\omega,t}^{S_i} \}_{t=1}^{N}$;
% raw gyroscope $\{ {y}_{\omega,t}^{S_i} \}_{t=1}^{N}$ for orientation propagation and centripetal term;
% initial quaternion $\hat{{q}}_{1|1}^{GS_i}$ and location vector $r_{1|1}^{S_i}$; covariance matrices $P_{1|1}$, $Q$, $R$.\\

\textbf{Output:} Relative orientation estimates $\hat{{q}}_{t|t}^{S_1S_2}$ and sensor location vector estimates $\hat{r}_{i, t|t}^{S_i}$, for $i=1,2$ and $t = 2, \hdots, N$.

\begin{algorithmic}[1]

\STATE Apply a zero-phase low-pass filter to $\{y_{a,t}^{S_i}\}_{t=1}^{N}$ and $\{y_{\omega,t}^{S_i}\}_{t=1}^{N}$ to obtain $\{\breve{y}_{a,t}^{S_i}\}_{t=1}^{N}$ and $\{\breve{y}_{\omega,t}^{S_i}\}_{t=1}^{N}$, and compute $\dot{\breve{y}}_{\omega,t}^{S_i}$ from $\breve{y}_{\omega,t}^{S_i}$ using 5-point numerical differentiation.

\FOR{{$t = 2, \ldots, N $} }
     \STATE \quad \textbf{Time update}
     \STATE \quad  $\hat{\eta}_{i,t|t-1}^G = 0, \quad i=1,2$,
     \STATE\quad $\hat{{q}}_{t|t-1}^{GS_i}
           = \hat{{q}}_{t-1|t-1}^{GS_i}
             \odot  \exp_q\!\left(\tfrac{T}{2}\,\bar{y}_{\omega,t-1}^{S_i}\right)$,
           \quad $i=1,2$,
           % \Comment{propagate with \emph{unfiltered} gyroscope (see~\eqref{eq:quat_prop})}
     \STATE \quad$\hat{{r}}_{i,t|t-1}^{S_i} = \hat{{r}}_{i,t-1|t-1}^{S_i}$,
           \quad $i=1,2$,
           % \Comment{constant-location prediction (see~\eqref{eq:r_rw})}
     \STATE \quad$P_{t|t-1} =  P_{t-1|t-1} + G_{t}\, Q\, G_{t}^\top$, with 
    $G_{t} = \mathrm{blkdiag}\!\left(
                            T\hat{{R}}_{t|t-1}^{GS_1},\;
                            T\hat{{R}}_{t|t-1}^{GS_2},\;
                            \mathcal{I}_3,\;
                            \mathcal{I}_3
                        \right)$.
    \STATE\quad \textbf{Measurement update}
    % \STATE \quad \quad Compute $\dot{\breve{{y}}}_{\omega,t}^{S_i}$ from $\breve{{y}}_{\omega,t}^{S_i}$ via 5-point numerical differentiation
    % % use low-pass filtered signal $\breve{{y}}_{a,t}^{S_i}$,and  .
    % % $\dot{\breve{y}}{\omega,t}^{S_i} = \Delta_5!\left(\breve{y}{\omega}^{S_i}\right)_t$
    %  \STATE \quad\quad Compute  ${a}_{jc,t}^{S_i}$  (see \eqref{eq:ajc-lpf})
    % \STATE \quad Compute $\dot{\breve{{y}}}_{\omega,t}^{S_i}$ from $\breve{{y}}_{\omega,t}^{S_i}$ via 5-point numerical differentiation and use
    
  % \STATE \quad Compute
  % $\hat{a}_{jc,t}^{S_i}=\breve{{y}}_{a,t}^{S_i} - \left(\left[{y}_{\omega,t}^{S_i} \times\right]^2 + \left[\dot{\breve{{y}}}_{\omega,t}
  % ^{S_i} \times\right]\right) \hat{r}_{i, t|t-1}^{S_i}$
  % \STATE \quad **and use it to compute** $H_t$ (see \eqref{eq:linearize_meas}) and the innovation ${\epsilon}_{\text{link},t} = \hat{R}
  % _{t|t-1}^{GS_1} \hat{a}_{jc,t}^{S_1} - \hat{R}_{t|t-1}^{GS_2} \hat{a}_{jc,t}^{S_2}$. Then update the state and state covariance as
    \STATE \quad Compute $\hat{a}_{jc,t}^{S_i}$ for $i=1,2$:
  \[
  \hat{a}_{jc,t}^{S_i}
  =
  \breve{{y}}_{a,t}^{S_i}
  -
  \left(\left[{y}_{\omega,t}^{S_i} \times\right]^2
  +
  \left[\dot{\breve{{y}}}_{\omega,t}^{S_i} \times\right]\right)
  \hat{r}_{i,t|t-1}^{S_i}.
  \]
  \STATE \quad Compute $H_t$ using \eqref{eq:linearize_meas}.
  \STATE \quad Compute the innovation ${\epsilon}_{\text{link},t} = \hat{R}_{t|t-1}^{GS_1} \hat{a}_{jc,t}^{S_1} - \hat{R}_{t|t-1}^{GS_2}
  \hat{a}_{jc,t}^{S_2}$. Then update the state and state covariance as
    % \STATE \quad  $\hat{a}_{jc,t}^{S_i}=\breve{{y}}_{a,t}^{S_i} - \left(\left[{y}_{\omega,t}^{S_i} \times\right]^2 + \left[\dot{\breve{{y}}}_{\omega,t}^{S_i} \times\right]\right) \hat{r}_{i, t|t-1}^{S_i}$
    % \STATE to compute $H_t$ (see \eqref{eq:linearize_meas}) and the innovation ${\epsilon}_{\text{link},t} = \hat{R}_{t|t-1}^{GS_1} \hat{a}_{jc,t}^{S_1} - \hat{R}_{t|t-1}^{GS_2} \hat{a}_{jc,t}^{S_2}$. Then update the state and state covariance as
     \STATE\quad $S_t = H_t\,P_{t|t-1}\,H_t^\top + R$,
           % \Comment{$S_t \in \mathbb{R}^{3\times 3}$ is the innovation covariance}
     \STATE\quad $K_t = P_{t|t-1}\,H_t^\top\,S_t^{-1}$,
     \STATE\quad $\hat{x}_{t|t} = \hat{x}_{t|t-1} + K_t\,{\epsilon}_{\text{link},t}$,
           % \Comment{state correction; prior mean is zero}
     \STATE \quad$P_{t|t} = P_{t|t-1} - K_t\,S_t\,K_t^\top$.
     \STATE\quad \textbf{Relinearize}
     \STATE\quad $\hat{{q}}_{t|t}^{GS_i}
           = \exp_q\!\left(\tfrac{\bar{\hat{\eta}}_{i,t|t}^G}{2}\right)
             \odot \hat{{q}}_{t|t-1}^{GS_i}$,
           \quad $i=1,2$,
    \STATE \quad $\hat{{q}}_{t|t}^{S_1S_2}=(\hat{{q}}_{t|t}^{GS_1})^{-1} \odot\hat{{q}}_{t|t}^{GS_2}$.
           \ENDFOR
\end{algorithmic}
\end{algorithm}

The measurement update (steps~8--15) uses the measurement model \eqref{eq:constraint_phys} to update $\hat{x}_{t|t-1}$.  The measurement update requires the joint-center acceleration  $a_{jc,t}^{S_i}$, which is computed from the sensor acceleration $a^{S_i}$, angular velocity $\omega^{S_i}$, and angular acceleration $\dot{\omega}^{S_i}$ in \eqref{eq:ajc}. We use the gyroscope measurement $y_{\omega,t}^{S_i}$ as an approximation of $\omega^{S_i}$. For the acceleration $a^{S_i}$, however, impacts, e.g. due to footsteps, can introduce high-frequency transients that are not well represented by the Gaussian measurement-noise model in \eqref{eq:acc_meas}, violating the assumptions of the EKF.   We therefore apply a zero-phase 4th-order Butterworth LPF to the accelerometer signal. We denote the output of this LPF by $\breve{y}_{a,t}^{S_i}$ and use it to approximate $a_t^{S_i}$ in~\eqref{eq:ajc}.
 % following the use of LPF for impulsive IMU measurement errors~\cite{Rong2024}.  
 Furthermore, differentiation of noisy signals is known to increase high-frequency noise. Thus, to determine $\dot{\omega}^{S_i}$, we first apply the LPF to $y_{\omega,t}^{S_i}$, to obtain the output denoted by $\breve{y}_{\omega,t}^{S_i}$, and then compute its 5-point numerical derivative $\dot{\breve{y}}_{\omega,t}^{S_i}$ as an approximation of $\dot{\omega}^{S_i}$.  Since the zero-phase LPF and the 5-point numerical differentiation use the measurement sequence, both are performed before the recursive EKF update.
% by 5-point numerical differentiation of the filtered sequence. We use $\breve{(\cdot)}$ to denote the LPF output: in \eqref{eq:ajc}, $a_t^{S_i}$  and $\dot{\omega}_t^{S_i}$ are replaced by $\breve{y}_{a,t}^{S_i}$ and  $\dot{\breve{y}}_{\omega,t}^{S_i}$, respectively. 
% Then, we compute the linearized measurement matrix $H_t \in \mathbb{R}^{3\times12}$ and the innovation ${\epsilon}_{\text{link},t}$  in step~11. 
% The innovation  ${\epsilon}_{\mathrm{link},t}$ is the difference between the  rigid-body constraint prediction and the measured IMU signals.
 The EKF needs the Jacobian $H_t \in \mathbb{R}^{3\times12}$ of the measurement model \eqref{eq:constraint_phys} around $\hat{x}_{t|t-1}$. 
 % The measurement matrix $H_t \in \mathbb{R}^{3\times12}$ is obtained by linearizing \eqref{eq:constraint_phys} around the state  $\hat{x}_{t|t-1}$. 
 Defining $\hat{a}^{G,i}_{jc,t} \triangleq \hat{R}_{t|t-1}^{GS_i} \hat{a}_{jc,t}^{S_i}$, we obtain 
\begin{equation}
H_t = \begin{pmatrix}
    [\hat{a}^{G,1}_{jc,t} \times] & -[\hat{a}^{G,2}_{jc,t} \times] &
    \hat{R}_{t|t-1}^{GS_1} C_t^{S_1} & -\hat{R}_{t|t-1}^{GS_2} C_t^{S_2}
\end{pmatrix} \label{eq:linearize_meas},
\end{equation}
where $C_t^{S_i} = - \left(\left[{y}_{\omega,t}^{S_i} \times\right]^2 + \left[\dot{\breve{{y}}}_{\omega,t}^{S_i} \times\right]\right)$.
The measurement noise covariance is 
% $R = {\Sigma}_{\text{link}} = \sigma_\text{link}^2\,\mathcal{I}_3$.
% $R = \Sigma_\text{link}\,\mathcal{I}_3$.
$R = \Sigma_\text{link}$.

%% file: chapters/sec_IV.tex
\section{Systematic simulation study}
\label{sec:sim_data}
% We first use simulated data to systematically evaluate Alg.1 under controlled conditions, and to study how estimation accuracy depends on motion intensity.
We systematically evaluate Alg.1 in simulation using the model in Section~\ref{sec:modeling} to characterize how estimation accuracy depends on motion intensity. We also compare Alg.1 with an adaptation of the method from~\cite{weygers2020drift} in which the sensor locations estimated using~\cite{seel2012joint} are replaced by the ground-truth $r_i^{S_i}$ (we refer to this method as Weygers-GT). This allows us to assess whether jointly estimating $r_i^{S_i}$ degrades relative orientation estimation accuracy. Since the simulation approximately matches the assumed Gaussian noise model, no LPF was used in Alg.1 for the simulation study. 

\subsection{Simulation settings}

%Each simulation run first sampled the geometry of two-segment rigid-body system. 
 For each Monte Carlo simulation,  both sensor location vectors $r_i^{S_i}$,  $i=1,2$, were assigned a fixed magnitude of $200$~mm and a random direction. The random direction was obtained by sampling from three independent zero-mean, unit-variance  Gaussians and scaling the resulting vector to unit length. The initial orientations $q_1^{GS_i}$, $i=1,2$, were generated with three angles sampled from $\mathcal{U}(0,2\pi)$.

Segment angular velocities were sampled as $\omega_t^{S_i}\sim\mathcal{N}\!\left(0,A_\omega^2 \mathcal{I}_3\right)$, with $ A_\omega\in\{0,0.01,0.1,0.5,1,1.35,3\}$~rad/s for $i =1,2$.     Given  $q_1^{GS_i}$ and $\omega_t^{S_i}$, ground-truth orientation  trajectories $q_t^{GS_i}$, $t = 1, \hdots , N$ were computed using \eqref{eq:quat_prop}, but replacing $y_{\omega,t}^{S_i}$ with $\omega_t^{S_i}$. The joint-center gravity-free linear acceleration $a_{jc,t}^{G,\mathrm{free}}$  was independently generated as a 3D  cubic spline with control points sampled from $\mathcal{N}(0,2^2)$ and scaled by  $A_{a_{jc}}\in\{0,0.05,0.1,0.5,1,2,10\}$~m/s$^2$.  Given $r_i^{S_i}$, $\omega_t^{S_i}$, and $a_{jc}^{G,\mathrm{free}}$, the  ground-truth sensor accelerations $a_t^{S_i}$, $i=1,2$, were obtained from  \eqref{eq:ajc}, with  
$a_{jc,t}^{S_i}=R_t^{S_iG}(a_{jc,t}^{G,\mathrm{free}}-g^G)$, $g^G=[0,0,9.82]^\top$~m/s$^2$ and  $\dot{\omega}_t^{S_i}$ computed from $\omega_t^{S_i}$ using 5-point numerical differentiation.

Accelerometer and gyroscope measurements $y_{a,t}^{S_i}$ and $y_{\omega,t}^{S_i}$, $i=1,2$, were simulated by adding white Gaussian noise with standard deviations $\sigma_{a,\mathrm{gen}}=0.02$~m/s$^2$ and  $\sigma_{\omega,\mathrm{gen}}=0.2^\circ$/s, respectively. Furthermore, sensor biases were added and sampled as $b_{a,\mathrm{gen}}\sim\mathcal{U}(-0.1,0.1)$~m/s$^2$ and  $b_{\omega,\mathrm{gen}}\sim\mathcal{U}(-0.5,0.5)^\circ$/s.

% Given  $q_1^{GS_i}$, and $\omega_t^{S_i}$,  we computed the ground-truth orientation  trajectories $q_t^{GS_i}$ and the corresponding accelerometer and gyroscope measurements $y_{a,t}^{S_i}$ and $y_{\omega,t}^{S_i}$ for $i=1,2$.
% We use the sampled geometry and excitation inputs to compute ground truth trajectories and simulated IMU measurements. 

% The ground truth  orientation trajectories were obtained from~\eqref{eq:quat_prop} using the sampled angular velocities and $q_1^{GS_i}$.
% The sensor accelerations were obtained from~\eqref{eq:ajc}, with gravity $g^G=[0,0,-9.82]^\top$~m/s$^2$ and angular acceleration $\dot{\omega}_t^{S_i}$ computed from the sampled angular velocity using 5-point numerical differentiation. 
% Simulated IMU measurements were obtained by adding white Gaussian noise and constant, per-run biases to  the ground truth accelerometer and gyroscope signals. Specifically, the generation noise levels are  $\sigma_{a,\mathrm{gen}}=0.02$~m/s$^2$ and  $\sigma_{\omega,\mathrm{gen}}=0.2^\circ$/s, while the biases are sampled as $b_{a,\mathrm{gen}}\sim\mathcal{U}(-0.1,0.1)$~m/s$^2$ and  $b_{\omega,\mathrm{gen}}\sim\mathcal{U}(-0.5,0.5)^\circ$/s.

We performed Monte Carlo simulations with 49 excitation scenarios, formed by  all combinations of the seven $A_{a_{jc}}$ levels and the seven $ A_\omega$ levels defined above. Each scenario included 10 independent 50-s simulations sampled at 100~Hz, resulting in a total of 490 simulations.
% The noise covariances used for $Q$ and $R$ in EKF are set as  $\sigma_\omega^2 = 0.01~\text{rad}^2\text{s}^{-2}$    $\sigma_r^2 = 1\times 10^{-9}$~m$^2$ in $Q$ and  $\sigma_\text{link}^2 = 1.225$~m$^2$s$^{-4}$ in $R$ in all experiments. $P_{1|1} = $

\subsection{Evaluation settings and evaluation method}
 % The performance of the proposed Alg.1 estimation was compared against the Weygers~\cite{weygers2020drift}. 
% We compare Alg.1 with a baseline, Weygers-GT. The original Weygers method~\cite{weygers2020drift} first estimates $r_i^{S_i}$ offline using Seel et al.~\cite{seel2012joint} and then fixes them during orientation filtering. Since simulation provides ground-truth $r_i^{S_i}$, Weygers-GT uses these true locations, giving a best-case fixed-location comparison for assessing whether online $r_i^{S_i}$ estimation in Alg.1 affects orientation accuracy.  
The orientation estimation accuracy is evaluated using the root mean squared angular distance (RMSD) between the estimated relative sensor orientation and the ground truth as $e_q = \sqrt{\frac{1}{N} \sum_{t=1}^{N} d_\angle (\hat{q}_t^{S_1S_2}, q_t^{S_1S_2})^2}$, where $d_\angle(\cdot,\cdot)$ is the angular distance between two quaternions, evaluated over the final 10~s of each simulation, allowing sufficient time for convergence. 
% The ground truth  orientation trajectories were obtained from~\eqref{eq:quat_prop} using the sampled angular velocities and $q_1^{GS_i}$. 
We define the sensor location error as the Euclidean distance between the ground truth and the N-th (final) estimate: $e_{r_i} = \|\hat{r}_{i,\mathrm{N}}^{S_i} - r_i^{S_i}\|_2$. 
% In the simulation study we report $e_{r_1}$ because the two sensors are symmetric by construction.

To quantify the motion of the system we use two metrics: the intensity of motion at the joint center $I_{a_{jc}} = \frac{1}{N} \sum_{t=1}^{N} \|a_{jc,t}^{G,\text{free}}\|_2$
and the intensity of relative rotation of two segments $I_{\omega} = \frac{1}{N} \sum_{t=1}^{N} \|{R}^{S_1 S_2}_t\,\omega_t^{S_2} - \omega_t^{S_1}\|_2$.

 In Alg.1, we choose the process noise covariance $Q$ in \eqref{eq:Q} as $\Sigma_\omega=\sigma_{\omega,\mathrm{gen}}^2\mathcal{I}_3$, matching the  gyroscope noise covariance used during data generation, and  $\Sigma_r=10^{-9}\mathcal{I}_3$, enforcing a nearly constant sensor location. The measurement noise covariance matrix $R$ is approximated as  $\sigma_{\mathrm{link}}^2 =2\left(\sigma_{a,\mathrm{gen}}^2+\sigma_{\omega,\mathrm{gen}}^2\right)$. 
For Alg.~1, we initialize $P_{1|1}=\mathcal{I}_{12}$, $\hat{q}_{1|1}^{GS_i}=[1,0,0,0]^\top$, and $\hat{r}_{1|1}^{S_i}=[1,1,1]^\top$~m. For Weygers-GT, we use the same $\hat{q}_{1|1}^{GS_i}$ and $P_{1|1}$ initialization as Alg.~1.

\subsection{Results}
\label{sec:sim_result}

\begin{figure}[h]
    \centering
    \includegraphics[width=1\linewidth]{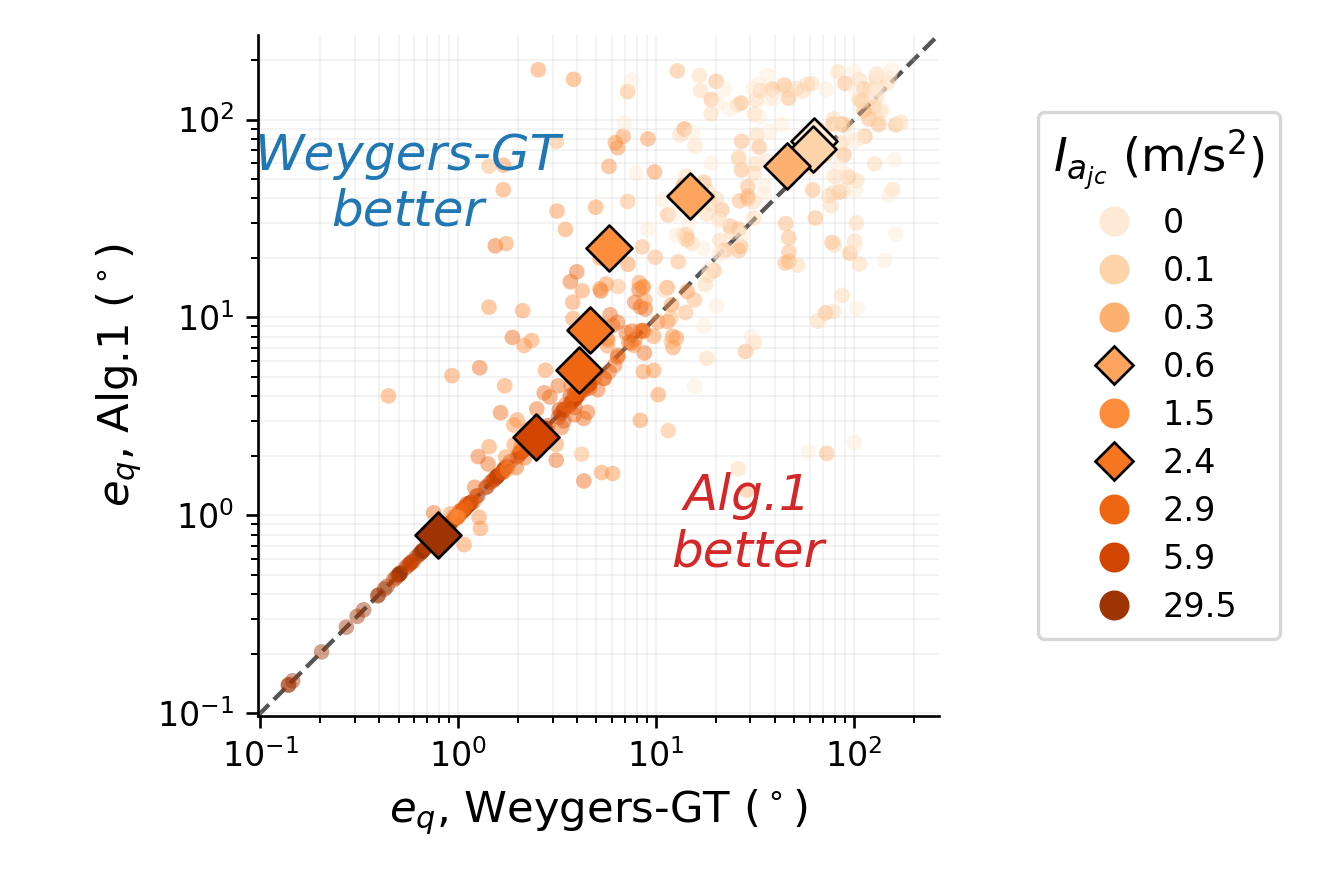}
    
    % \includegraphics[width=0.75\linewidth]{plots/paper_figure/oom.png}
    % The zoom-in sub-plot shows additional $I_{a_{jc}}$ levels (the hollow diamond markers) around $I_{a_{jc}}=1.5$~m/s$^2$.
    \caption{
	    Alg.1 vs.\ Weygers-GT orientation error $e_q$. Small dots are individual simulations for different intensity levels and diamond markers show the mean at each $I_{a_{jc}}$ level encoded by color. The dashed line indicates equal performance between the two methods. To illustrate the trend around $I_{a_{jc}}=$1.5~m/s$^2$, we additionally show the mean $e_{q}$ at $I_{a_{jc}}=$0.6 and $2.4$~m/s$^2$ as diamond markers, omitting individual simulations for clarity.
    % $e_q^{\text{RMSD}}$ 
    }
    \label{fig:sim-aim1}
\end{figure}

Fig.~\ref{fig:sim-aim1} compares the orientation estimation accuracy of Alg.1  and Weygers-GT. At the three lowest intensity levels at the joint center ($I_{a_{jc}}=0$, $0.1$, and $0.3$~m/s$^2$), orientation estimation errors are large for both methods because there is insufficient excitation.  For both methods, the orientation estimation error $e_q$ decreases as $I_{a_{jc}}$ increases.
% As can be seen in Fig.~\ref{fig:sim-aim1}, the  largest performance difference occurs at the intermediate level $I_{a_{jc}}=1.5\,$~m/s$^2$, where the mean $e_q$ is $22.4^\circ$ for Alg.1 and $5.8^\circ$ for Weygers-GT. 
% The zoom-in view further shows additional $I_{a_{jc}}$ levels that Alg.~1 also has larger errors at neighboring intermediate levels between, while the errors of the two methods become similar toward both ends of this range.
% This difference suggests that jointly estimating the sensor locations affects orientation estimation when the joint-center acceleration intensity is still limited. As $I_{a_{jc}}$ increases towards $I_{a_{jc}}=2.9$~m/s$^2$, this difference decreases, and Alg.~1 approaches the orientation accuracy obtained with known sensor locations.
As shown in Fig.~\ref{fig:sim-aim1}, the largest difference occurs at $I_{a_{jc}}=1.5$~m/s$^2$, where the mean $e_q$ is $22.4^\circ$ for Alg.~1 and $5.8^\circ$ for Weygers-GT. 
Results at two additional $I_{a_{jc}}$ levels, $0.6$ and $2.4$~m/s$^2$, show larger errors for Alg.~1 around $1.5$~m/s$^2$ and a decreasing difference toward $2.9$~m/s$^2$.
This suggests that jointly estimating the sensor locations affects orientation estimation at intermediate $I_{a_{jc}}$, while Alg.~1 approaches the accuracy of Weygers-GT as $I_{a_{jc}}$ increases.
From $I_{a_{jc}}=2.9$~m/s$^2$ onward, Alg.1 and Weygers-GT achieve similar orientation accuracy, with mean $e_{q}$ values of approximately $5.4^\circ$  or less. The mean $e_q$ of Alg.~1 further decreases to $2.5^\circ$ and $0.8^\circ$ at $I_{a_{jc}}=5.9$ and $29.5$~m/s$^2$, respectively, becoming comparable to Weygers-GT, indicating that online sensor location estimation does not degrade relative orientation estimation at sufficiently high $I_{a_{jc}}$. 
It is worth noting that the ground-truth sensor locations $r_i^{S_i}$ are only available in simulation.

  \begin{figure}[ht]
  \centering
  \includegraphics[width=\linewidth]{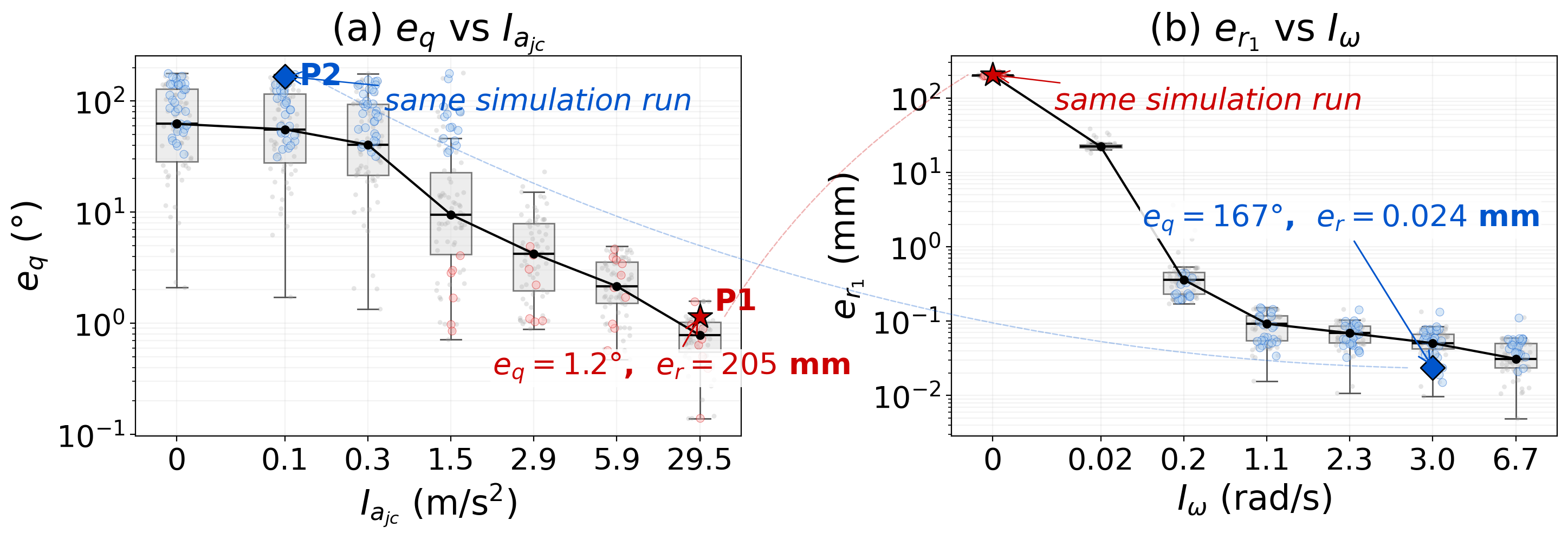}
      \caption{Alg.1 errors vs.\ motion intensity in simulation for  (a)~$e_q$ vs.\ $I_{a_{jc}}$ and (b)~$e_{r_1}$ vs.\ $I_\omega$.  Boxes show interquartile ranges with median lines.
  Red markers denote low-$e_q$/high-$e_{r_1}$ simulations ($e_q<5^\circ$,
  $e_{r_1}>100\,\mathrm{mm}$), and blue markers denote low-$e_{r_1}$/
  high-$e_q$ simulations ($e_{r_1}<0.5\,\mathrm{mm}$, $e_q>30^\circ$). The star and diamond markers represent simulations P1 and P2 as two examples.}
  \label{fig:sim_2x2}
\end{figure}

  % Given this relationship between joint-center acceleration intensity and orientation error, we next examine whether sensor location error follows the same excitation source or instead follows relative angular velocity.

To further analyze Alg.1, we next examine the dependence of its estimation errors $e_q$ and $e_{r_i}$ on the intensity of motion at the joint center, $I_{a_{jc}}$, and the intensity of relative rotation, $I_\omega$.
Since the errors for the two sensors are nearly identical, we report $e_{r_1}$ only. 
As shown in Fig.~\ref{fig:sim_2x2}, $e_q$ decreases with increasing $I_{a_{jc}}$, whereas $e_{r_1}$ decreases with increasing $I_\omega$. We quantify the strength of the relationships between $e_q$ and $I_{a_{jc}}$, and between $e_{r_1}$ and $I_\omega$, using the coefficient of determination, denoted as $R^2$, obtained by fitting linear regressions to the log-transformed quantities. Zero-excitation  levels are excluded from the regressions because their logarithms are  undefined.
For $e_q$, $R^2=0.71$ with $I_{a_{jc}}$ and $R^2=1.5\times10^{-2}$ with $I_\omega$. For $e_{r_1}$, $R^2=0.90$ with $I_\omega$ and $R^2=4.1\times10^{-5}$ with $I_{a_{jc}}$. The contrast in $R^2$ values indicates that orientation accuracy is primarily associated with joint-center acceleration intensity $I_{a_{jc}}$, whereas the  accuracy of the estimated sensor location vectors $\hat{r}_i^{S_i}$ is primarily associated with relative angular-velocity intensity $I_\omega$. 
\begin{table*}[!t]
  \caption{
    Relative orientation error $e_q$ in degrees after a 50~s convergence window.  Values are means across all subjects. Overall denotes the mean across all joints, subjects, and motion types. The smallest (best) value in each column is bold.}
  \label{tab:main_results}
  \centering
  \small
  \begin{tabular}{ll cccr cccr r}
    \toprule
    & & \multicolumn{4}{c}{Walking ($n_\text{subject}$=4)} & \multicolumn{4}{c}{Complex Tasks ($n_\text{subject}$=3)} & \\
    \cmidrule(lr){3-6}\cmidrule(lr){7-10}
    Method & $r_i^{S_i}$ source & Hip & Knee & Ankle & Avg & Hip & Knee & Ankle & Avg & Overall \\
    \midrule
    % \multicolumn{11}{l}{\textit{Primary comparison}} \\[2pt]
    Alg.1 & IMU (Online)
        & \textbf{2.55} & \textbf{1.99} & \textbf{1.72} & \textbf{2.09}
        & \textbf{4.20} & \textbf{2.91} & \textbf{2.79} & \textbf{3.30}
        & \textbf{2.61} \\
    Weygers \cite{weygers2020drift}   & IMU (pre-cal. \cite{seel2012joint})
        & 5.36 & 11.03 & 7.40 & 7.93
        & 19.32 & 6.95 & 14.85 & 13.71
        & 10.40 \\
    \midrule
    % \multicolumn{11}{l}{\textit{Ablation: $r_i^{S_i}$  Marker (pre-cal.)\ vs online }} \\[2pt]
    Weygers-Mrk  & Marker (pre-cal.)
        & 5.69 & 4.70 & 4.69 & 5.03
        & 12.31 & 8.71 & 10.86 & 10.63
        & 7.43 \\
    Weygers-Mrk-LPF & Marker (pre-cal.)
        & 2.82 & 2.63 & 2.20 & 2.55
        & 4.29 & 3.88 & 3.29 & 3.82
        & 3.09 \\
    Alg.1-no-LPF         & IMU (Online)
        & 4.71 & 3.42 & 5.22 & 4.45
        & 11.24 & 6.87 & 12.01 & 10.04
        & 6.85 \\
    % \midrule
    % \multicolumn{11}{l}{\textit{Ablation: LPF alternatives}} \\[2pt]
    Alg.1-NIS       &  IMU (Online)
        & 2.93 & 2.33 & 2.45 & 2.57
        & 5.14 & 7.13 & 3.82 & 5.36
        & 3.77 \\
    Alg.1-SM         &  IMU (Online)
        & 4.58 & 3.58 & 5.88 & 4.68
        & 12.61 & 6.91 & 12.67 & 10.73
        & 7.27 \\
    \bottomrule
    % \multicolumn{11}{l}{\footnotesiz Across all subjects, the per-subject standard divination of $\bar{e}_q$ ranges from $0.3°$–$0.4°$ (ankle/knee, walking) to $1.0°$ (hip, walking).  } \\
    % \multicolumn{11}{l}{\footnotesiz Complex tasks standard divination is below $1.9°$ for all joints.   } \\
    % markdowns/kf_manon_eval_results.csv
   
  \end{tabular}
\end{table*}

% The individual simulation provide the same interpretation.  Among the highlighted simulations in Fig.~\ref{fig:sim_2x2}, 30 have accurate orientation but large sensor location error (red markers), whereas 77 show the opposite behavior (blue markers). 
 % For  example, P1 has $e_q=2.0^\circ$ and $e_{r_1}=201\,\mathrm{mm}$, whereas P2 has  $e_{r_1}=0.046\,\mathrm{mm}$ and $e_q=174^\circ$. These examples further  confirm that accurate orientation estimation does not necessarily imply accurate sensor location estimation, and vice versa.

%% file: chapters/sec_V_A.tex
% \subsection{Dataset and Experimental Setup}
% \label{sec:dataset}
% \section{Experiments}
% \label{sec:experiments}

\section{Experiments}
\label{sec:experiments}

We evaluate Alg.1 using experimental data from the OpenSense dataset~\cite{al2022opensense}, considering subjects with a complete lower-body seven-IMU configuration, i.e., with sensors on the pelvis, upper legs, lower legs, and feet. The experimental data consist of walking motion from subjects 02, 03, 04, and 07, and complex tasks from subjects 02, 03, and 04, including sitting, stair ascent and descent, side-stepping, and running.
Inertial data were recorded with Xsens MTw IMUs at 100~Hz, and  optical motion-capture ground truth was recorded with a Vicon system at 100~Hz. Each dataset was synchronized by cross-correlating the IMU and marker data using the right femur angular velocity magnitudes, and truncated to 600~s.

Marker-based sensor locations ${r}_{i,\text{Mrk}}^{S_i}$ were obtained as illustrated in Fig.~\ref{fig:sensor_diagram} (right), using the centroid of the  optical markers placed around the IMU   and the marker-based joint center. For the latter, the processed joint center provided by OpenSense is used for the hip, while the closest skin-mounted marker is used for the knee and ankle. The resulting sensor locations are time-averaged for each dataset to obtain $\bar{r}_{i,\text{Mrk}}^{S_i}$.

%We use marker-based sensor location ${r}_{i,\text{Mrk}}^{S_i}$ as the Marker. 
%These are determined by time-averaging the vector from the center of the optical markers placed around the IMU (see Fig.~\ref{fig:sensor_diagram} (b)) to the skin-mounted joint marker, expressed in the sensor frame. 
Let $O_i$ denote the optical-marker-cluster frame attached to IMU $i$. We calculate the Vicon-based relative orientation as
$ q^{O_1O_2} =  \left(q^{GO_1}\right)^{-1}\odot q^{GO_2}$ and align the IMU and optical-marker frames, $S_i$ and $O_i$, using Theorem~4.2 from \cite{hol2011sensor},  following \cite{weygers2020drift}, to obtain the ground truth $q_t^{S_1S_2}$. 

The orientation estimation error is again evaluated using~$e_q$, see Section~\ref{sec:sim_data}. We compute  $e_q$ over the data after a conservative 50~s convergence window for all methods to ensure convergence across all methods. Since the sensor location is assumed constant, we define the per-sensor location error as $e_{r_i}=\|\hat{r}_{i,\mathrm{N}}^{S_i}-\bar{r}_{i,\mathrm{Mrk}}^{S_i}\|_2$ for $i\in\{1,2\}$ for each adjacent sensor pair around each joint. To quantify the movement intensity, we use the same metrics as defined in Section~\ref{sec:sim_data}, with signals replaced by measurements: $I_{a_{jc}} = \frac{1}{N} \sum_{t=1}^{N} \|a_{jc,t}^{G,\text{free}}\|_2$, where $a_{jc,t}^{G,\text{free}}$ is approximated by twice differentiating the marker-based joint center location and $I_{\omega} = \frac{1}{N} \sum_{t=1}^{N} \|{R}^{S_1 S_2}_t\,y_{\omega,t}^{S_2} - y_{\omega,t}^{S_1}\|_2$, where $y_{\omega,t}^{S_i}$ are IMU gyroscope measurements and ${R}^{S_1 S_2}_t$ is derived from the Vicon ground-truth $q_t^{S_1S_2}$.

Following~\cite{kok2017using}, if the first sample satisfies $0.6\|g^G\| < \|y_{a,t}^{S_i}\| < 1.4\|g^G\|$, the initial orientation $\hat{q}_{1|1}^{GS_i}$ is computed from this sample by assuming that the first accelerometer sample provides inclination information and by initializing the heading to be zero. 
The cut-off frequency of the LPF was chosen as $f_c = 6~\mathrm{Hz}$, consistent with common filtering practices for human gait analysis \cite{winter2009biomechanics}. The noise covariances are set as $\sigma_\omega^2 = 0.01~\mathrm{rad^2\,s^{-2}}$ and $\sigma_\mathrm{link}^2 = 1.225~\mathrm{m^2\,s^{-4}}$, with $\sigma_r^2 = \sigma_\mathrm{link}^2 \cdot 10^{-9}$. The initial covariance and sensor location vector values are chosen identical to Section~\ref{sec:sim_data}: $P_{1|1} = \mathcal{I}_{12}$, $\hat{r}_{1|1}^{S_i} = [1,1,1]^\top~\mathrm{m}$.

\subsection{Relative orientation estimation accuracy}
\label{sec:orientation}

% used by  Weygers~\cite{weygers2020drift}, 

For the walking and complex-task datasets, we evaluate our proposed method, Alg.1, against the method from~\cite{weygers2020drift} (from here on referred to as Weygers) which uses the offline sensor location procedure~\cite{seel2012joint}, and with five variants of Alg.1 and Weygers:  Weygers-Mrk, Weygers-Mrk-LPF, Alg.1-no-LPF, Alg.1-NIS, and
Alg.1-SM. Weygers requires prior knowledge of the sensor  locations, obtained using the offline procedure from~\cite{seel2012joint}. We also evaluate Weygers using marker-based sensor locations, referred to as Weygers-Mrk, to study the effect of potentially erroneous sensor location estimates on the performance of Weygers. Alg.1 extends Weygers not only by jointly estimating the  relative orientation and sensor locations, but also by applying an LPF to the IMU signals. To analyze the contribution of the LPF, we compare Alg.1 with Alg.1-no-LPF, which removes the LPF, and with Alg.1-NIS, which replaces the LPF with normalized innovation squared (NIS)-based update rejection. Finally, we compare to Alg.1-SM, where the state is augmented with the angular velocity, similar to \cite{lorenz2025minjointtracker}. Here, $\dot{\omega}$ is obtained from the augmented state rather than using 5-point numerical differentiation as in Alg.1.

 Table~\ref{tab:main_results} reports the orientation error $e_q$ for both the  walking and complex-task trials. Alg.1 achieves an overall error of  $2.61^\circ$, a 75\% reduction relative to Weygers ($10.40^\circ$), partly because Weygers uses less accurate sensor locations obtained with the  offline procedure from \cite{seel2012joint}. However, Alg.1 also outperforms Weygers-Mrk ($7.43^\circ$) by 65\%, despite its use of marker-based Vicon sensor locations. The LPF is an important reason for this: Weygers-Mrk-LPF improves over Weygers-Mrk by 58\%, and Alg.1 improves over Alg.1-no-LPF by 62\%, with consistent gains across all  joints and motion types. Without the LPF, Alg.1-no-LPF ($6.85^\circ$)  performs similarly to Weygers-Mrk. With online sensor location estimation and the  LPF, Alg.1 ($2.61^\circ$) outperforms Weygers-Mrk-LPF ($3.09^\circ$),  despite the latter using marker-based sensor locations obtained through an offline calibration procedure requiring an optical motion-capture system. The lower $e_q$ of Alg.1 may partly arise because $\bar{r}_{i,\text{Mrk}}^{S_i}$ are only approximations of ${r}_i^{S_i}$.

 % and flexibility in time-varying $\hat{r}^{S_i}_{i,t}$ rather than using fixed ${r}_{i,\text{Mrk}}^{S_i}$.
 
 % These could be explained by out-of-joint marker bias in $r_{mrk}$ and also flexibility in $\hat{r}_i^{S_i}$ compared with fixed sensor location used in Weygers-Mrk-LPF.
 % while achieving comparable or better orientation accuracy on the evaluated experimental data.
 
 On average, across the walking trials and both sides, $I_{a_{jc}}$  increases from the hip to the knee and ankle ($1.98$, $3.88$, and $5.21$~m/s$^2$, respectively), whereas $e_q$ decreases ($2.55^\circ$, $1.99^\circ$, and $1.72^\circ$). At the individual-trial level, considering the two 3D joints, the hip exhibits lower $I_{a_{jc}}$ and  higher $e_q$ than the ankle in every subject-side walking trial. This implies that, in line with the simulation results, higher joint-center acceleration  provides more information through the kinematic constraint and improves relative orientation accuracy.

  % Table~\ref{tab:main_results} show that LPF is the main factor behind the improvement on real data: Weygers-Mrk-LPF ($3.09^\circ$) improves over Weygers-Mrk ($7.43^\circ$) by 58\%, and Alg.1 ($2.61^\circ$) improves over Alg.1-no-LPF ($6.85^\circ$) by 62\%, with consistent gains across all joints and motion types. Notably, without LPF, Alg.1-no-LPF ($6.85^\circ$) performs comparably to the Weygers-Mrk ($7.43^\circ$) with marker-based pre-calibration. With LPF, Alg.1 ($2.61^\circ$) outperforms the Weygers-Mrk-LPF ($3.09^\circ$). Given that Alg.1 does not require an offline marker-based calibration procedure, these results suggest that the proposed method estimating $r_i^{S_i}$ online provides a practical alternative while achieving comparable or improved accuracy in the evaluated dataset.
% To motivate LPF as a physically necessary step rather than an empirical tuning choice, we examine the \emph{Normalised Innovation Squared} (NIS).
  % In conclusion, Table~\ref{tab:main_results} indicates that online sensor location estimation with Alg.1 is a practical alternative to offline calibration method, achieving comparable or better orientation accuracy on the evaluated experimental data, as jointly estimating the sensor locations alone does not by itself guarantee improved orientation accuracy. The lower error of Alg.1 ($2.61^\circ$) compared with Alg.1-no-LPF ($6.85^\circ$) suggests that mitigating outliers is important for Alg.1's improved performance from Weygers. To examine this effect, we compare the NIS of Alg.1 and Alg.1-no-LPF.  

\subsection{Effect of low-pass filtering}
  \label{sec:lpf_effect}

The results above show that online sensor location estimation can replace offline calibration on  experimental data. The improvement in accuracy does not come from online sensor location estimation alone. We therefore examine why the LPF improves the estimation and whether alternative approaches can result in a similar improvement.

% In the link-constraint measurement model in \eqref{eq:constraint_phys}, the measurement noise is assumed Gaussian. In real IMU measurements,  this assumption can be violated by outliers. Although 100~Hz sampling captures the main lower-body motion, foot-strike impacts can introduce high-frequency acceleration transients that are not well represented by the Gaussian measurement-noise model~\cite{Mitschke2017,Fan2025}. Furthermore, numerical differentiation of measured angular velocity amplifies high-frequency noise~\cite{Winter1990}. Together, these effects can make the link-constraint innovation deviate from the assumed measurement model and degrade estimation accuracy.

  % To further study this effect, we compute the NIS for  Alg.1 and Alg.1-no-LPF.
\begin{figure}[t]
    \centering
    \includegraphics[width=1\linewidth]{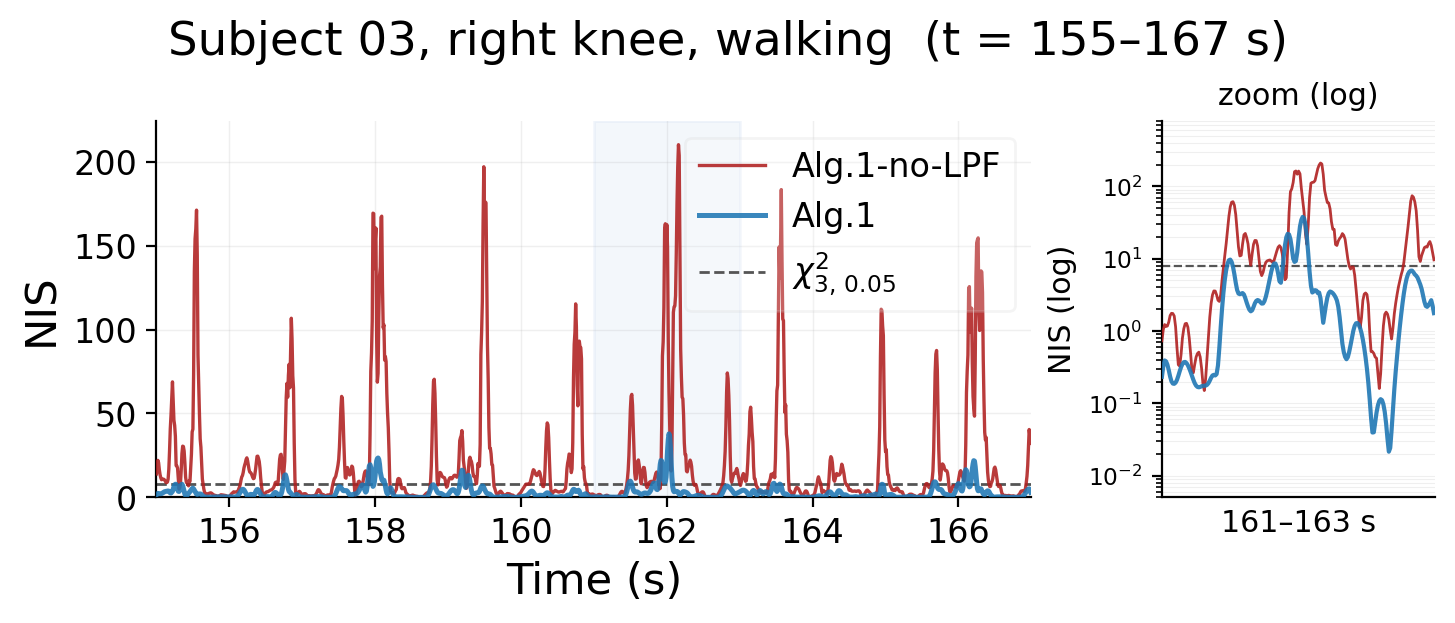}
    \caption{%
    NIS over a 12-second walking segment (Subject~03, right knee).   The zoomed window shows approximately one gait cycle.
       %  NIS over a 12-second walking segment
       %  (Subject~03, right knee).
       %  Alg.1-no-LPF exhibits periodic peaks at the foot-strike frequency ($\approx$0.5--1~Hz).
       % Alg.1 exhibits reduced peak magnitudes and NIS generally below the $\chi^2$ threshold.
       }
    \label{fig:nis}
\end{figure}
Fig.~\ref{fig:nis} shows the NIS over a 12-second walking segment (Subject~03, right knee). For Alg.1-no-LPF, the NIS exhibits sharp periodic  spikes at the foot-strike frequency, where acceleration transients and noise  amplified by numerical differentiation cause the innovation $\epsilon_{\mathrm{link}}$ to deviate from its assumed Gaussian model, resulting in NIS values outside the expected $\chi^2$ range. Alg.1  suppresses these spikes, and its NIS remains mostly within the expected $\chi^2$ range.

The large NIS values at specific time instants suggest that outlier rejection could provide an alternative to the LPF. We therefore analyze Alg.1-NIS, which skips the measurement update when $\text{NIS}_t={\epsilon}_{\text{link},t}^{\top}S_t^{-1}\epsilon_{\text{link},t}$ 
exceeds the $\chi^2$  threshold $\chi_{3,0.95}^2=7.81$.
We additionally consider Alg.1-SM, which uses angular velocity state augmentation to avoid numerical differentiation. Alg.1-NIS achieves an overall orientation error of $3.77^\circ$,  worse only than Alg.1 and Weygers-Mrk-LPF.
However, fixed-threshold outlier rejection can result in valid measurements being repeatedly rejected and may require an additional recovery mechanism~\cite{berman2014outliers}.
In contrast, Alg.1 achieves a lower error of $2.61^\circ$, without these limitations. 
Alg.1-SM achieves $7.27^\circ$, comparable to Alg.1-no-LPF ($6.85^\circ$), indicating that state augmentation  does not resolve the model inconsistency. 
These results support the use of the LPF over NIS-based outlier rejection (Alg.1-NIS) or angular velocity state augmentation (Alg.1-SM).

\subsection{Online sensor location estimation}

\label{sec:r1r2}

 We analyze the  sensor location estimates to study the differences across the different methods, and further examine differences across joints and their relation to the relative angular velocity intensity identified in simulation. We  compare the locations estimated  by Alg.1 and Alg.1-no-LPF with the offline-calibrated sensor locations from \cite{seel2012joint} used by Weygers.
 % ~\cite{weygers2020drift}, which are obtained using the offline procedure described by Seel et al.~\cite{seel2012joint}.  

For each evaluated joint, we consider the two IMUs attached to the adjacent body segments, with corresponding sensor location vectors $r_1^{S_1}$ and $r_2^{S_2}$. The sensor location error $e_{r_i}$ is computed for both $i=1,2$ and averaged.

\begin{figure}[ht]
\centering
% Replace with \includegraphics{filename} later
% \rule{\columnwidth}{3in} 
    \includegraphics[width=1\linewidth]{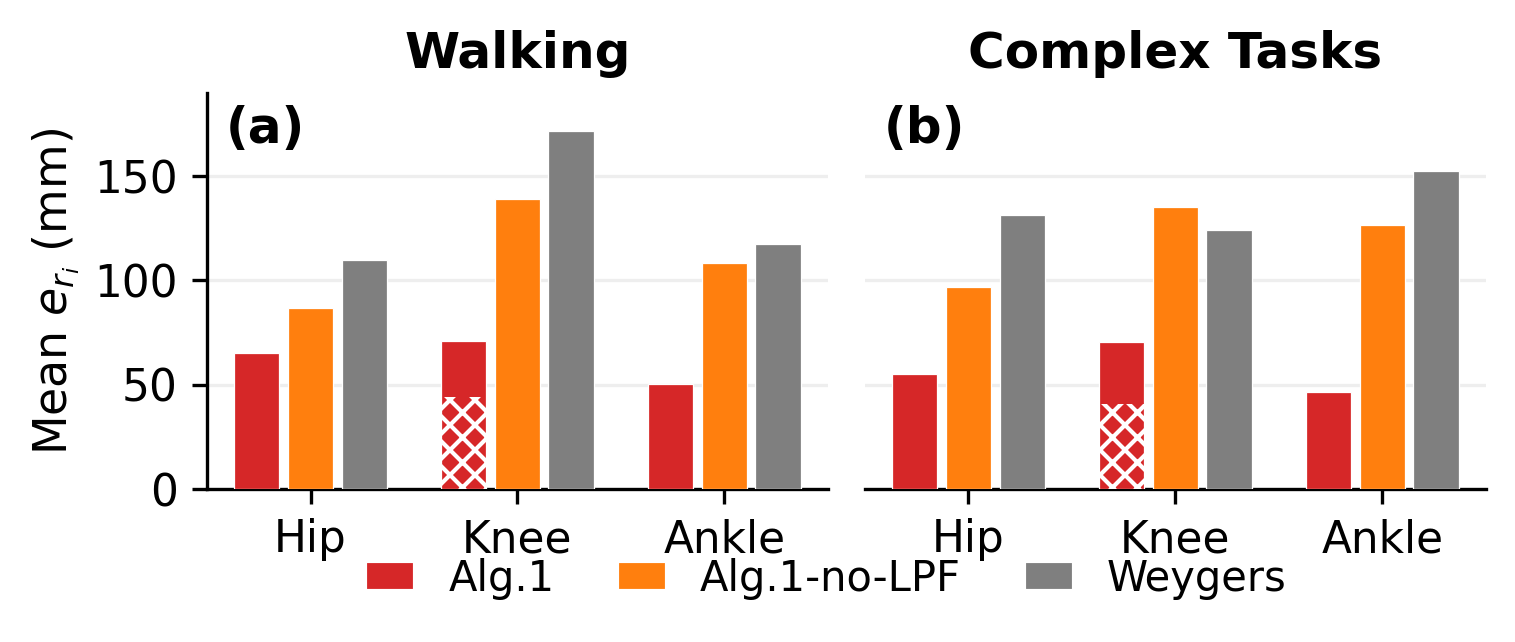}
\caption{Sensor location error $e_{r_i}$ averaged over sensors $i = 1, 2$, left/right sides, and subjects, for each joint and for both walking and complex tasks. The cross-hatched region on the Alg.1 knee bar indicates the error of the observable perpendicular component as described by~\cite{olsson2017experimental}.}
 % averaged over both adjacent sensors ($i=1,2$), left/right sides, and subjects.
\label{fig:r1r2}
\end{figure}

Fig.~\ref{fig:r1r2} shows that Alg.1 achieves the lowest mean sensor  location error, $60$~mm, compared with $115$~mm for Alg.1-no-LPF and  $134$~mm for the fixed locations used by Weygers. This  difference is most visible at the ankle, where both Alg.1-no-LPF and the  fixed locations used by Weygers exceed $100$~mm. Weygers and Weygers-Mrk use the same orientation filter, but Weygers relies on less accurate $r_{i,\mathrm{Seel}}^{S_i}$, which partly explains its larger orientation error relative to Weygers-Mrk in  Table~\ref{tab:main_results}.  Compared with Alg.1-no-LPF, Alg.1 reduces the mean sensor location error from  $115$~mm to $60$~mm, showing that LPF improves not only relative orientation  estimation but also online sensor location estimation. 
% Please note that since the reference $r_{mrk}^{S_i}$ are still not truth joint location as external marker not the joint center, $e_{r_i}$ serves as a difference

 % These sensor location errors also help interpret the orientation results in  Table~\ref{tab:main_results}. Weygers and Weygers-Mrk use the same orientation   filter, but Weygers relies on less accurate $r_{i,\mathrm{Seel}}^{S_i}$,  which partly explains its larger orientation error relative to Weygers-Mrk.   Compared with Alg.1-no-LPF, Alg.1 reduces the mean sensor location error from  $115$~mm to $60$~mm, showing that LPF improves not only relative orientation  estimation but also online sensor location estimation. 

 % We next analyze the joint- and motion-dependent errors of Alg.1. 
Since the simulation results indicate that sensor location accuracy depends on motion excitation, we examine how the estimation errors differ across joints and motion types.  The ankle  has the lowest mean error ($50$~mm for walking and $47$~mm for complex tasks), followed by the hip ($65$~mm and $55$~mm), whereas the knee has the largest error ($\sim71$~mm for both motion types). The larger knee error is partly  structural: under hinge-like motion, the component of the sensor location vector along the rotation axis is only partially observable~\cite{olsson2017experimental}.  Evaluating only the observable component perpendicular to this axis reduces the knee error to $44$~mm for walking and $41$~mm for complex tasks, as indicated by the cross-hatched region in Fig.~\ref{fig:r1r2}. The corresponding overall mean error decreases from $60$~mm to $51$~mm. 
 
To examine whether these joint- and motion-dependent differences are associated with the relative angular velocity intensity identified in simulation, we compare the errors with $I_\omega$. The knee error is nearly unchanged between walking and complex tasks  ($70.9$ versus $70.5$~mm), whereas the hip and ankle errors are lower during  complex tasks by $10.2$~mm and $3.6$~mm, respectively. For the two 3D joints, the ankle has lower error than the hip in both  walking ($50.2$ versus $65.4$~mm) and complex tasks ($46.6$ versus $55.2$~mm), and also has higher mean $I_\omega$ in both motion types. This is consistent  with the simulation result that greater relative angular velocity provides  more information for estimating sensor locations, although differences in joint geometry and marker-based uncertainty may also contribute. 
% \vspace{2cm}

\subsection{Additional  complex motion analysis}
\label{sec:case_study}
\begin{figure}[!h]
  \centering
  \includegraphics[width=1\linewidth,  trim={14 0 10 0},
    clip]{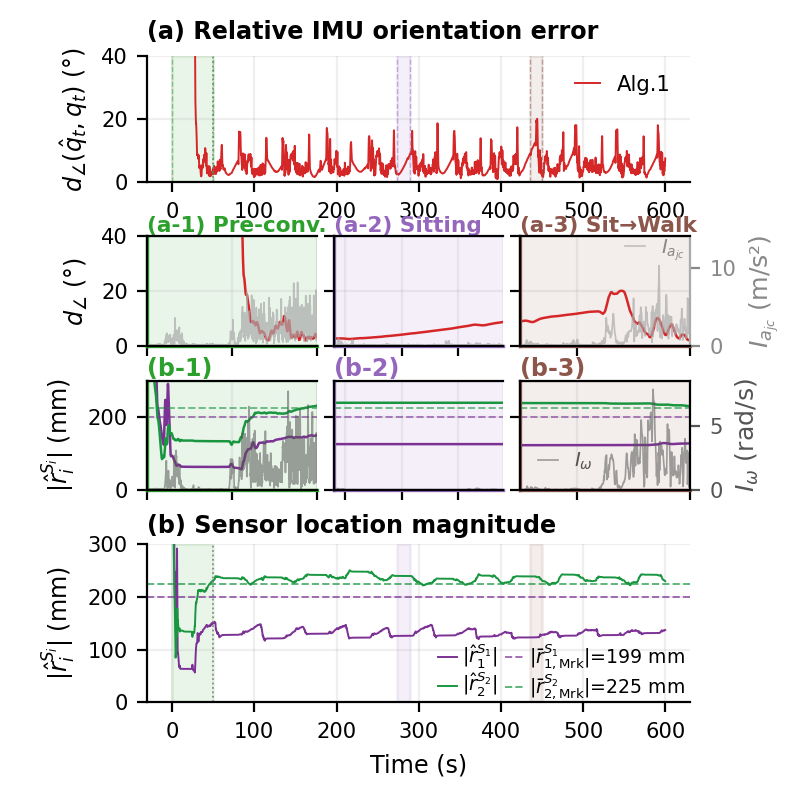}
  \caption{Subject~04, right hip, complex motion. (a) Relative orientation error over the full trial, with the shaded intervals shown in the zoomed views. (b) Magnitudes of the estimated sensor location over the full trial, with the same zoomed views. Gray curves show the intensities. 
  % $|\hat{r}_i^{S_i}|$ 
  \label{fig:case_study}}
\end{figure}

To characterize Alg.1 beyond trial-averaged performance, we analyze its  behavior over time under varying excitation levels. We examine a representative Subject~04 right-hip trial during the complex tasks  (Fig.~\ref{fig:case_study}), which contains intervals of both low and high   motion intensity. This allows us to evaluate how the relative orientation and  sensor location estimates evolve as the excitation varies.
For this, we define the time-varying intensities
$I_{a_{jc},t}=\left\|a_{jc,t}^{G,\mathrm{free}}\right\|_2$ and $I_{\omega,t}=\left\|R_t^{S_1S_2}y_{\omega,t}^{S_2}-y_{\omega,t}^{S_1}\right\|_2$.

Fig.~\ref{fig:case_study}(a) shows the relative orientation error $e_q$ over the full trial, with the zoomed windows in  Fig.~\ref{fig:case_study}(a-1)--(a-3) highlighting the transition  between low and high $I_{a_{jc},t}$.  The orientation error stays large during the first 25\,s until sufficient $I_{a_{jc},t}$ appears (Fig.~\ref{fig:case_study}(a-1)), consistent with the intensity dependence established in  Section~\ref{sec:sim_data}. During sitting (Fig.~\ref{fig:case_study}(a-2)), $ I_{a_{jc},t}\approx 0$  renders the kinematic constraint uninformative, resulting in slow orientation drift.  Notably, this drift stops within seconds  when sufficient $I_{a_{jc},t}$ resumes (Fig.~\ref{fig:case_study}(a-3)),   suggesting that the method effectively recovers after periods of low intensity.
% The error accumulates when the subject is stationary, but recovers within 

% Fig.~\ref{fig:case_study}(b) shows the estimated sensor location  magnitudes $|\hat{r}_i^{S_i}|$ over time. The estimates converge once $I_{\omega,t}$ increases, and remain more stable for the remainder of the trial. Once converged, when $I_{\omega,t}$ is low, $\hat{r}_i^{S_i}$ does not change due to the low process noise, see Fig.~\ref{fig:case_study}(b-2). When $I_{\omega,t}$ increases again, only minor fluctuations are observed, see Fig.~\ref{fig:case_study}(b-3).
% Overall, the observed behavior is consistent with the simulation trends for $I_{a_{jc}}$ and $I_\omega$. 

% Interestingly, the marker based ri is not constant as \ref{fig:r1r2} or Weygers assumed. This might also explain the better performance of alg.1 than weygers-mrk-lpf in section \ref{sec:experiments}. Further research can be done to model the dynamic in $r_i$ on different segments.
Fig.~\ref{fig:case_study}(b) shows the estimated sensor location magnitudes $|\hat{r}_{i,t}^{S_i}|$ over time, together with the constant marker-based reference $|\bar{r}_{i,\mathrm{Mrk}}^{S_i}|$. The estimates settle as $I_{\omega,t}$ increases, remain nearly unchanged during low $I_{\omega,t}$ due to the small process noise (b-2), and exhibit modest fluctuations when $I_{\omega,t}$ increases again (b-3). Unlike Weygers, which assumes fixed sensor locations, our method models $r_i^{S_i}$ as approximately constant but allows small temporal variations through process noise (see \eqref{eq:r_rw}). Whether this flexibility improves relative orientation accuracy remains to be investigated.
% Fig.~\ref{fig:case_study}(b) shows the estimated sensor location magnitudes $|\hat{r}_i^{S_i}|$ over time. The estimates settle as $I_{\omega,t}$ increases, remain nearly unchanged during low $I_{\omega,t}$ because of the small process noise (b-2), and exhibit modest fluctuations when $I_{\omega,t}$ increases again (b-3).  Interestingly, the magnitudes of marker-based sensor location vary over time, in contrast to the constant sensor location assumption in Weygers, Weygers-Mrk, and Weygers-Mrk-LPF. Although our proposed method  models  $r_i^{S_i}$ as approximately constant (see \eqref{eq:r_rw}), its process noise permits small temporal variations. Whether allowing time-varying locations improves relative orientation accuracy, remains to be investigated in the future.

% Overall, the results are consistent with the simulation: relative orientation accuracy depends mainly on $I_{a_{jc},t}$, while sensor location accuracy depends on $I_{\omega,t}$.

%% file: chapters/conclusion.tex
\section{Conclusion}
We developed a calibration-free, magnetometer-free EKF that jointly estimates relative IMU orientation and sensor location for human motion estimation. To improve reliability on real IMU measurements, our method mitigates high-frequency measurement outliers caused by foot-strike impacts and numerical differentiation using an LPF.
 Simulation results show that online sensor location estimation does not degrade orientation accuracy under sufficient joint center acceleration. Furthermore, simulations show that orientation and sensor location accuracy primarily depend on joint center acceleration and relative angular velocity, respectively.
Experimental results demonstrate that the proposed method achieves an overall relative orientation error of $2.61^\circ$, which is 75\% lower than the existing method using IMU-based offline sensor location calibration~\cite{weygers2020drift}. The evaluation covers both walking and complex tasks, extending validation beyond the walking motions considered in previous studies. Future work includes extending relative IMU orientation estimation to anatomical joint angles through automatic sensor-to-segment calibration.

\section*{Acknowledgment}

% This work was funded by the Sensor AI Lab, under the AI Labs program of Delft University of Technology, and by the Convergence Human Mobility Center.
This work was funded by the Sensor AI Lab under the AI Labs program of Delft University of Technology (TU Delft) and by Convergence Health \& Technology (Convergence Human Mobility Centre), a flagship initiative funded by the Convergence Alliance (TU Delft, Erasmus MC, Erasmus University Rotterdam). OpenAI's ChatGPT was used for language suggestions, refinement of author-written code, and generation of lower-limb sketch elements in the graphical abstract and Fig.~2. All AI-assisted material was reviewed and verified.